\documentclass[11pt]{article}

\usepackage[preprint]{acl}

\usepackage{svg}
\usepackage{booktabs}
\usepackage{tabularx}
\usepackage{array}
\usepackage{colortbl}
\usepackage{makecell}
\usepackage{multirow}
\usepackage{supertabular}
\usepackage{amsmath}
\usepackage{amssymb}
\usepackage{placeins}
\usepackage{needspace}
\usepackage{tcolorbox}
\usepackage{subcaption}
\usepackage{times}
\usepackage{latexsym}
\usepackage{footnote}
\makesavenoteenv{figure*}

\usepackage{lipsum}
\usepackage{comment}

\usepackage[T1]{fontenc}
\usepackage[utf8]{inputenc}

\usepackage{microtype}

\usepackage{inconsolata}

\usepackage{graphicx}

\newcounter{fdpromptno}[section]
\newlength\fdpromptparindent
\newenvironment{fdprompt}[1]
{
  \setlength{\fdpromptparindent}{\the\parindent}
  \setlength{\parindent}{0pt}
  \refstepcounter{fdpromptno}
  \par\medskip
  \noindent
  \begin{tcolorbox}[left=1pt,right=1pt]
  \textsc{Template \thesection.\thefdpromptno: #1}\\
  \small
  \ttfamily
}
{
  \end{tcolorbox}
  \setlength{\parindent}{\fdpromptparindent}
  \medskip
}

\title{Don't Let Me Ask for It: LLMs Show Deficiencies in Active Multi-Turn Information Acquisition for Abductive Inference}

\author{First Author \\
  Affiliation / Address line 1 \\
  Affiliation / Address line 2 \\
  Affiliation / Address line 3 \\
  \texttt{email@domain} \\\And
  Second Author \\
  Affiliation / Address line 1 \\
  Affiliation / Address line 2 \\
  Affiliation / Address line 3 \\
  \texttt{email@domain} \\}

\author{%
Shahrukh Mohiuddin${^\mathbf{1}}$\thanks{Both authors equally contributed.}, Chalamalasetti Kranti${^\mathbf{1}}$\footnotemark[1], Sherzod Hakimov${^\mathbf{1}}$, David Schlangen${^\mathbf{1,2}}$\\$^{\mathbf{1}}$Computational Linguistics, Department of Linguistics\\
University of Potsdam, Germany\\
$^{\mathbf{2}}$German Research Center for Artificial Intelligence (DFKI), Berlin, Germany\\
{\texttt{\{shahrukh.mohiuddin, kranti.chalamalasetti, sherzod.hakimov, david.schlangen\}@uni-potsdam.de}}
}  

\begin{document}
\maketitle
\begin{abstract}
Abductive reasoning requires forming hypotheses that explain observed evidence and revising them as new evidence becomes available. While large language models (LLMs) are often evaluated on whether they solve abductive reasoning tasks correctly, less is known about how they acquire evidence, update their hypotheses, and decide when to stop. We introduce Alien Abduction game, an interactive probe for studying these behaviours under different interaction modes. The modes vary in whether evidence is provided upfront or across turns, and whether queries are selected by the model or examples are provided by the oracle. Across models, providing evidence upfront leads to higher success rates than distributing it across turns. In multi-turn settings, some models commit before using the available evidence, while others exhaust the turn budget without converging. Models also achieve higher success rates when examples are provided by the oracle than when they select their own queries, although their final hypotheses are more consistent with the evidence they selected. These findings suggest that models may form hypotheses that fit self-selected evidence without sufficiently distinguishing them from alternatives, and may struggle to validate and refine their hypotheses or determine when to stop.
\end{abstract}

\section{Introduction}
{\small\textit{One fact leads to another—so we continue. Does the next fit in with that? A merveille! Good! We can proceed. This next little fact—no! Ah, that is curious! There is something missing—a link in the chain that is not there. We examine. We search. And that little curious fact, that possibly paltry little detail that will not tally, we put it here!}
\\ -- Agatha Christie, in The Mysterious Affair at Styles~\footnote{\url{https://www.gutenberg.org/cache/epub/863/pg863-images.html\#chap04}}.
\par}
\vspace{2pt}

\begin{figure}[t]
    \centering
    \vspace{-0.3cm}
    \includegraphics[width=\columnwidth]{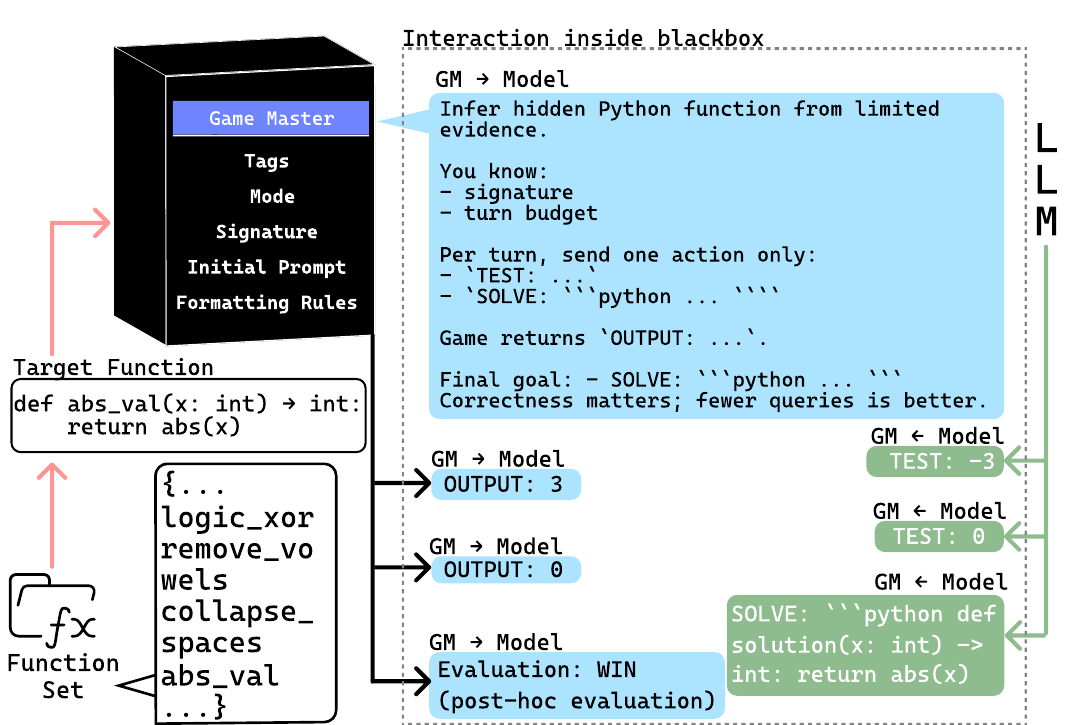}
\caption{Overview of \textit{Alien Abduction}. The Game Master hides a target Python function $f$ and interacts with the LLM through a black-box protocol. In the Active-Output mode shown here, the model proposes test inputs, receives the corresponding outputs, and eventually submits a final hypothesis as Python code. The submitted hypothesis is then evaluated after submission against the hidden target function on held-out test cases.}
    \label{fig:main_image}
    \vspace{-0.3cm}    
\end{figure}

\noindent
The ability to recognize an underlying rule from specific observations is a hallmark of intelligence: it has long been studied in cognitive science~\citep{wason1960failure, bruner2017study}, is a standard component of human IQ tests and is central to scientific discovery. In accounts of human inquiry dating back to Peirce~\citep{peirce1934collected}, such discovery is not a single inference step but a cycle: forming a candidate explanation from observations (abduction), deriving what that explanation predicts (deduction), and revising it against new evidence (induction) \citep{he2025idea,yin2026oracle}. LLMs now perform strongly on many reasoning benchmarks~\citep{NEURIPS2022_18abbeef, DBLP:journals/corr/abs-2407-21783, NEURIPS2024_53384f20}, yet these benchmarks typically present all information upfront and evaluate a single, non-interactive answer, often on data that may have appeared in their training corpora~\citep{balloccu-etal-2024-leak, ICLR2024_46e624c2, DBLP:journals/corr/abs-2502-14425}. They therefore cannot tell us whether a model can: decide what evidence to gather, form an explanation of it, and test that explanation before committing. This capability matters in practice, as LLMs are increasingly deployed as agents that must handle unfamiliar environments, tools, and APIs through interaction rather than specifications.

Several lines of work address parts of this problem. Interactive program synthesis benchmarks let agents query a hidden function and refine candidate programs~\citep{wei2025codearc,lee2025syntra}; inductive reasoning benchmarks study rule discovery from provided examples or active queries~\citep{honovich2023instruction,sun2024itd,DBLP:conf/nips/YanLLYFSDC25,chari2026wild}, while code benchmarks evaluate generation from explicit specifications or reasoning over visible programs~\citep{chen2021evaluating,austin2021program,gu2024cruxeval,DBLP:conf/acl/XuC00LHHC025}. Black-box reasoning environments extend hidden-rule discovery to broader domains~\citep{he2025idea,yin2026oracle,chen2025physgym}

However, these settings do not jointly vary who selects the evidence and what form the feedback takes over. Existing comparisons of active and passive evidence collection use exact outputs only~\citep{geng2025reliable}, leaving the role of binary verdicts and disconfirming evidence less clear. Moreover, evaluations focus on final-answer correctness without testing whether a model's intermediate hypotheses remain consistent with the evidence it has collected. As a result, the effects of self-directed exploration, negative evidence, and hypothesis grounding remain underexplored.

We introduce \textit{Alien Abduction}, a multi-turn game, in which the model must reconstruct a hidden Python function from only its signature and a limited turn budget; submissions are executed in a sandbox against held-out test cases (Figure~\ref{fig:main_image}). Four interactive game modes vary who controls the evidence (active vs. passive) and its form (exact outputs vs. membership verdicts on proposed input--output pairs); two single-turn modes serve as non-interactive baselines. This design measures the effect of self-directed exploration, tests whether models seek disconfirming rather than only confirming evidence, and scores submissions for consistency with the observed evidence. Targets span five domains: numbers, number pairs, strings, lists, and Boolean logic.

Our contributions are: (a) A benchmark game for black-box function induction with six interaction modes varying evidence control (active vs. passive) and evidence type (exact outputs vs. membership verdicts); (b) A suite of 50 automatically generated and validated target functions across five domains; (c) A sandboxed evaluation protocol with task success, turn-budget use and turn-level hypothesis analysis; and (d) An empirical study across commercial and open-weight models.

\section{Related Work}

\paragraph{Code generation and code understanding.}
HumanEval \citep{chen2021evaluating} and MBPP \citep{austin2021program} assess whether models can produce Python functions from natural-language descriptions, validated through unit tests; later work explored architectures and training for program synthesis \citep{nijkamp2023codegen2,zheng2023survey}. By contrast, CRUXEval \citep{gu2024cruxeval} and its multilingual extension CRUXEval-X \citep{DBLP:conf/acl/XuC00LHHC025} test reasoning about visible programs via input and output prediction. In both families the task is fully specified: the behavior is described or the code is shown. In Alien Abduction, neither is available; the behavior must be inferred before code can be written.

\paragraph{Inductive reasoning from examples.}
Instruction Induction \citep{honovich2023instruction} asks models to describe a hidden rule from input--output examples in natural language, ItD \citep{sun2024itd} improves inductive ability by leveraging deduction on list-function tasks, and MIR-Bench \citep{DBLP:conf/nips/YanLLYFSDC25} scales the setting to many-shot pattern recognition over hundreds of examples. In all of these, the evidence set is fixed. Moving beyond static observation, \citet{chari2026wild} study active concept learning with predefined query-selection policies (expected-information-gain vs. confirmation-style positive tests); the querying strategy is still not chosen by the model.

\paragraph{Interactive black-box reasoning environments.}
\citet{geng2025reliable} compare passive observation with active intervention when LLMs reverse-engineer programs, formal languages, and equations. In RULEARN \citep{he2025idea}, agents learn hidden rules through cycles of abduction, deduction, and induction; ORACLE \citep{yin2026oracle} extends black-box interaction to code, circuits, physical systems, encryption, and games. AR-Bench tests whether models can ask useful questions under incomplete information \citep{zhou2025arbench}, and PHYSGYM evaluates experiment design under controlled levels of prior knowledge \citep{chen2025physgym}. These benchmarks show that models often struggle to choose informative actions and to adapt after feedback; none, however, scores the final hypothesis for consistency with the evidence collected during the episode, or penalizes inefficient use of the interaction budget.

\paragraph{Interactive program synthesis.}
Closest to our setting, CodeARC \citep{wei2025codearc} evaluates agents that query a hidden target function and refine candidate Python implementations against a differential testing oracle, while SYNTRA \citep{lee2025syntra} selects informative test inputs to eliminate competing program hypotheses; \citet{surana2026iprog} add human feedback and verification. A complementary line uses black-box access for explanation rather than reconstruction \citep{cifka2023context,singh2023explaining}.

Alien Abduction combines these threads in a single controlled setting: varying evidence control and feedback form over identical hidden functions allows the effect of self-directed exploration to be estimated; membership-verdict feedback tests whether models seek disconfirming evidence; and an internal-consistency check separates inferring a wrong rule from ignoring one's own evidence.

\section{Methodology}
\subsection{Task Formulation}
\label{sec:task_formulation}
Let $f : X \rightarrow Y$ denote a hidden target function with graph
\[F_f = \{(x, y) \in X \times Y \mid y = f(x)\}.\]
The player never observes the source code of $f$. Instead, a Game Master with oracle access to $f$ mediates all interaction: it reveals evidence about $F_f$ according to a fixed protocol and verifies proposed solutions. Since $X$ may be multivariate, an input $x$ consists of one or more values.

An episode is parameterized by the function signature (argument and return types), an interaction mode, and a turn budget $T$. In each turn $t \leq T$, the player performs exactly one action: it either requests evidence, as permitted by the mode, or submits a candidate implementation $\hat{f}$ as Python code via the \texttt{SOLVE} action, which ends the episode. Alongside each action, the player reports its current hypothesis, a state indicating whether it is probing, confirming, or uncertain, its rationale for selecting the query, and a confidence score. These fields are recorded for analysis but are not processed by the Game Master and do not affect its response. An episode counts as solved if and only if $\hat{f}$ agrees with $f$ on a set of held-out test inputs; this set is constructed per instance and never revealed to the player, and agreement is checked by executing $\hat{f}$ in an ephemeral container (Section~\ref{sec:containerization}). An episode that exhausts its budget without a submission counts as failed.

Figure~\ref{fig:main_image} shows an episode in the \textit{Active-Output} mode: given the signature \texttt{(x: int) -> int}, the player probes the black box with inputs such as $-3$ and $0$, observes the outputs $3$ and $0$, and submits an implementation of the absolute-value function. The challenge is therefore not code generation from a specification, but inference of the behavioral rule from limited evidence. Each episode thereby instantiates the reasoning cycle from the introduction: the player forms a candidate rule from its observations (abduction), derives which probe would test it (deduction), and revises the rule as new evidence arrives (induction).

\subsection{Game Modes}
\label{sec:game_variants}

\begin{table*}[t]
\centering
\scriptsize
\setlength{\tabcolsep}{4pt}
\renewcommand{\arraystretch}{1.2}
\begin{tabular}{@{}l l l l l@{}}
\toprule
\textbf{Game mode} & \textbf{Control} & \textbf{Feedback} & \textbf{Player action} & \textbf{Example exchange} \\
\midrule
Active-Output (AO) & active & output & \texttt{TEST: <input>} & \texttt{TEST: -8} $\rightarrow$ \texttt{OUTPUT: 8} \\
\arrayrulecolor{black!30}\midrule
Active-Verdict (AV) & active & membership & \makecell[l]{\texttt{TEST: <input>,}\\\texttt{\ \ <output>}} & \makecell[l]{\texttt{TEST: 2, 2} $\rightarrow$ \texttt{OUTPUT: True}\\\texttt{TEST: 2, 1} $\rightarrow$ \texttt{OUTPUT: False}} \\
\arrayrulecolor{black!30}\midrule
Passive-Output (PO) & passive & output & \texttt{NEXT} & \texttt{NEXT} $\rightarrow$ \texttt{OUTPUT: (-2, 2)} \\
\arrayrulecolor{black!30}\midrule
Passive-Verdict (PV) & passive & membership & \texttt{NEXT} & \makecell[l]{\texttt{NEXT} $\rightarrow$ \texttt{OUTPUT: ((-2, 2), True)}\\\texttt{NEXT} $\rightarrow$ \texttt{OUTPUT: ((10, 9), False)}} \\
\arrayrulecolor{black!30}\midrule
\makecell[l]{Single-Turn-\\Output (STO)} & single-turn & output & \texttt{SOLVE} only & 10 examples shown upfront \\
\arrayrulecolor{black!30}\midrule
\makecell[l]{Single-Turn-\\Verdict (STV)} & single-turn & membership & \texttt{SOLVE} only & 10 labeled pairs shown upfront \\
\arrayrulecolor{black}\bottomrule
\end{tabular}
\caption{The six interaction modes of Alien Abduction, factorized by evidence control (who selects the probes) and feedback form (exact outputs vs. binary membership verdicts). In every mode the player can end the episode at any turn with \texttt{SOLVE:} followed by a candidate Python implementation; the single-turn modes permit only this action after a fixed batch of ten evidence items is shown. All example exchanges assume the hidden target function \texttt{absolute\_value(x: int) -> int}.}
\label{tab:alien_abduction_modes}
\vspace{-0.3cm}
\end{table*}

The six modes, summarized in Table~\ref{tab:alien_abduction_modes}, vary two properties of the evidence about $F_f$: who controls its selection (active vs. passive) and what form the feedback takes (exact outputs vs. binary membership verdicts).

In the two active modes, the player selects its own probes. \textit{Active-Output} (AO) returns the exact output $f(x)$ for a queried input $x$, yielding positive evidence only. \textit{Active-Verdict} (AV) instead returns whether a proposed pair $(x,y)$ lies in $F_f$: the player trades exact outputs for the ability to test, and potentially falsify, its own hypotheses, since both positive and negative evidence can result.

In the two passive modes, the Game Master controls the evidence stream. \textit{Passive-Output} (PO) reveals one valid pair from $F_f$ per request, reducing the problem to induction from externally provided positive evidence; \textit{Passive-Verdict} (PV) reveals candidate pairs with true/false labels, providing contrastive evidence without player control.

The two single-turn modes, \textit{Single-Turn-Output} (STO) and \textit{Single-Turn-Verdict} (STV), present a fixed batch of evidence upfront and permit only the \texttt{SOLVE} action. Since their evidence is still passively provided, they are the natural baselines for the passive modes: STO vs.\ PO isolates the effect of sequential interaction, while PO vs.\ AO isolates control over probe selection. Jointly, the six modes disentangle strategic probe selection, induction from positive evidence, induction from mixed positive and negative evidence, and hypothesis formation under a fixed interaction budget.
\section{Experiment Setup}

\subsection{Target Functions and Test Cases}
\label{sec:target_functions}

\begin{table}[t]
\centering
\scriptsize
\setlength{\tabcolsep}{3pt}
\renewcommand{\arraystretch}{1.1}
\begin{tabular}{@{}l l c@{}}
\toprule
\textbf{Domain} & \textbf{Signature} & \textbf{\#Fns} \\
\midrule
Number & \texttt{int $\rightarrow$ int} & 10 \\
Number Pairs & \texttt{(int, int) $\rightarrow$ int} & 10 \\
String & \texttt{str $\rightarrow$ str} & 10 \\
List & \texttt{List[int] $\rightarrow$ int} & 10 \\
Logic & \texttt{(bool, bool) $\rightarrow$ bool} & 10 \\
\bottomrule
\end{tabular}
\caption{The five domains of Alien Abduction, their function signatures, and the number of sampled target functions per domain. The complete list of target functions is given in Table~\ref{tab:alien_abduction_functions} in Appendix~\ref{app:functions}.}
\label{tab:alien_abduction_domains}
\vspace{-0.3cm}
\end{table}

The benchmark is built on five domains, each defined by a fixed function signature (Table~\ref{tab:alien_abduction_domains}): numbers, number pairs, strings, lists, and Boolean logic. We deliberately restrict the targets to primitive functions: short, pure, deterministic transformations that use only the standard library. Such functions keep the hypothesis space tractable, ensure that failures reflect evidence-gathering and rule inference rather than programming difficulty, and still admit many distinct behaviors per signature.

Target functions are generated rather than hand-picked. For each signature, we prompt GPT5.4 to produce 100 candidate functions that must match the signature exactly, be pure and deterministic, use no imports, and remain short. Each candidate is automatically validated: its code must parse, define exactly one function with the expected parameters, and execute without error on sample inputs. Because the targets are drawn from the generator model's own output distribution, they are by construction functions that an LLM can express, so failure to identify them cannot be attributed to complex behavior. From the validated pool we randomly sample (with a fixed seed) 10 functions per signature, yielding 50 targets. The same 50 targets are used across the modes, so differences between modes cannot be confounded by target difficulty.

Each target is paired with 100 test cases. Inputs are drawn from type-aware pools that mix edge cases (e.g., zero, negatives, empty strings and lists) with random values, and the output for every input is computed by executing the target function. These test cases serve two roles: they provide the evidence revealed in the passive and single-turn modes, and they act as the held-out suite against which every submitted hypothesis is verified.

\subsection{Models}
We evaluate four models: \texttt{GPT5.4}\footnote{\url{https://openai.com/index/introducing-gpt-5-4/}}, \texttt{GPT5.4-mini}\footnote{\url{https://openai.com/index/introducing-gpt-5-4-mini-and-nano/}}, and \texttt{Mistral-Large-3}~\citep{DBLP:journals/corr/abs-2601-08584} as commercial systems, and \texttt{Qwen3.6-35B-A3B}~\citep{qwen36_35b_a3b} as an open-weight model. Since GPT5.4 is also the generator of the target functions, its results additionally indicate how well a model recovers functions drawn from its own distribution.

\subsection{Metrics}
We report three metrics. \textit{Success rate} measures correctness: an episode scores 1 if the submitted hypothesis passes all held-out test cases of the hidden function, and 0 otherwise. Averaged over episodes and reported on a 0--1 scale, it is aggregated per mode, per domain, and per model.

\textit{Turn Budget Use} (TBU) measures the proportion of the available interaction budget consumed before an episode ends. For the interactive modes,
\[
\text{TBU} = (\frac{n- 1}{T})
\]
where $T$ is the maximum turn budget and $n$ is the number of turns used. A score of $1$ indicates that the episode uses the entire turn budget, while lower scores indicate earlier termination. For the single-turn modes, the turn budget is fixed at one and is therefore fully used by definition.

We measure \textit{hypothesis retrodiction accuracy} (HRA) to assess the consistency of a model's current hypothesis with the accumulated evidence. At turn $t$, we compute

\begin{equation*}
\mathrm{HRA}_t =
\frac{N_t^{\mathrm{matched}}}{N_t}
\end{equation*}

where $N_t^{\mathrm{matched}}$ is the number of observed evidences reproduced correctly by the hypothesis and $N_t$ is the total number observed. A score of $1$ indicates consistency with all available evidence. An agent that reliably tracks its belief state should maintain a score of $1$, since its reported hypothesis should not contradict evidence received so far. We report HRA across turns and at the final turn to examine how models revise and validate their hypotheses before committing.

TBU and HRA provide complementary views of the interaction. TBU captures how much of the available turn budget is used, while HRA captures whether the reported hypothesis remains consistent with the accumulated evidence.

\subsection{Implementation}
\label{sec:containerization}
Alien Abduction~\footnote{The source code and test instances will be released upon the paper’s acceptance} is implemented in the clembench framework \citep{chalamalasetti-etal-2023-clembench}, which provides the Game Master loop, prompt handling, and scoring infrastructure. Each interactive episode has a turn budget of $T = 15$; the single-turn modes allow a single \texttt{SOLVE} action. In total, each model plays the 50 targets in all six modes, i.e., 300 episodes per model. Models are accessed through their respective APIs with default decoding parameters. At each turn, we log the \textit{selected action}, \textit{query and response}, \textit{current hypothesis}, \textit{interaction state}, \textit{query rationale}, \textit{confidence score}, and whether the model \textit{response was parsed successfully}. These traces are used to compute the turn-level metrics and support the qualitative analysis. Every submitted solution is executed in a native Python~3 environment inside an ephemeral sandbox container that is destroyed after the run; only the test results are returned to the Game Master, which then decides whether the episode is won or lost.

\begin{figure*}[t]
\centering
  \vspace{-0.3cm} 
  \includegraphics[width=0.88\textwidth]{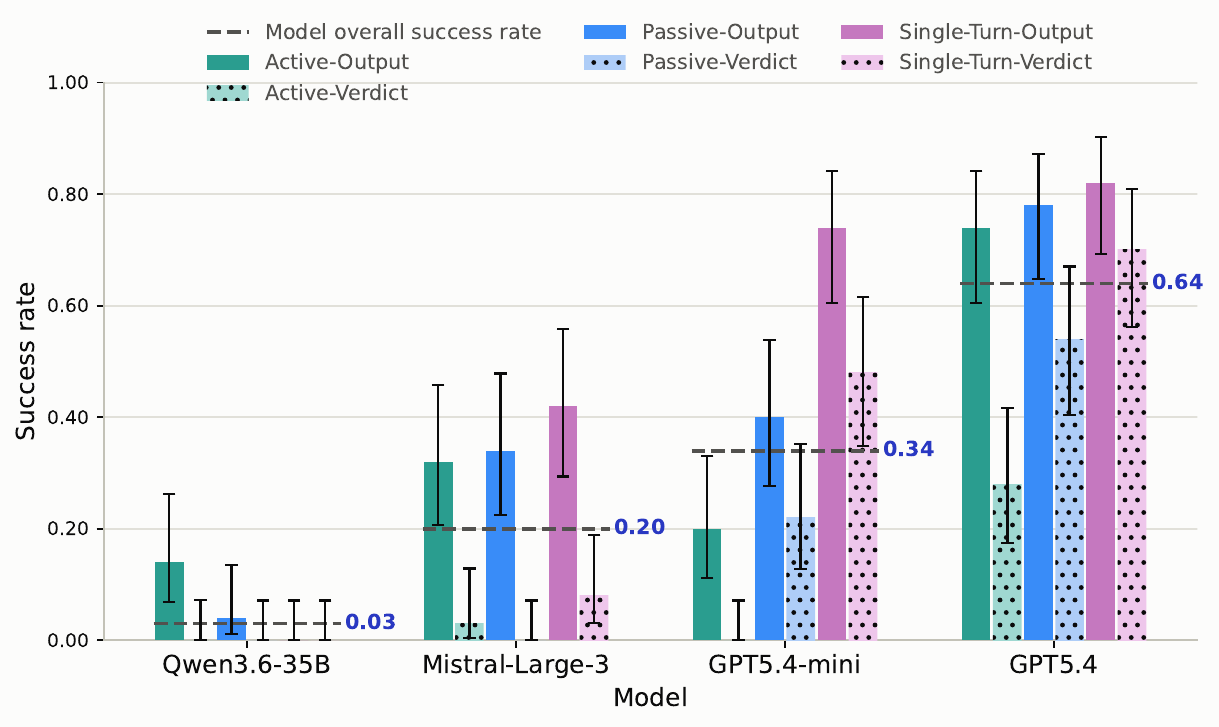}
  \caption{Success rates across interaction modes for each model, with 95\% confidence intervals. Dashed horizontal lines indicate the overall success rate for the model. A missing bar indicates that zero success for that mode.}
  \label{fig:overallsuccessrate}
  \vspace{-0.3cm}  
\end{figure*}

\begin{figure*}[t]
\centering
  \includegraphics[width=0.88\textwidth]{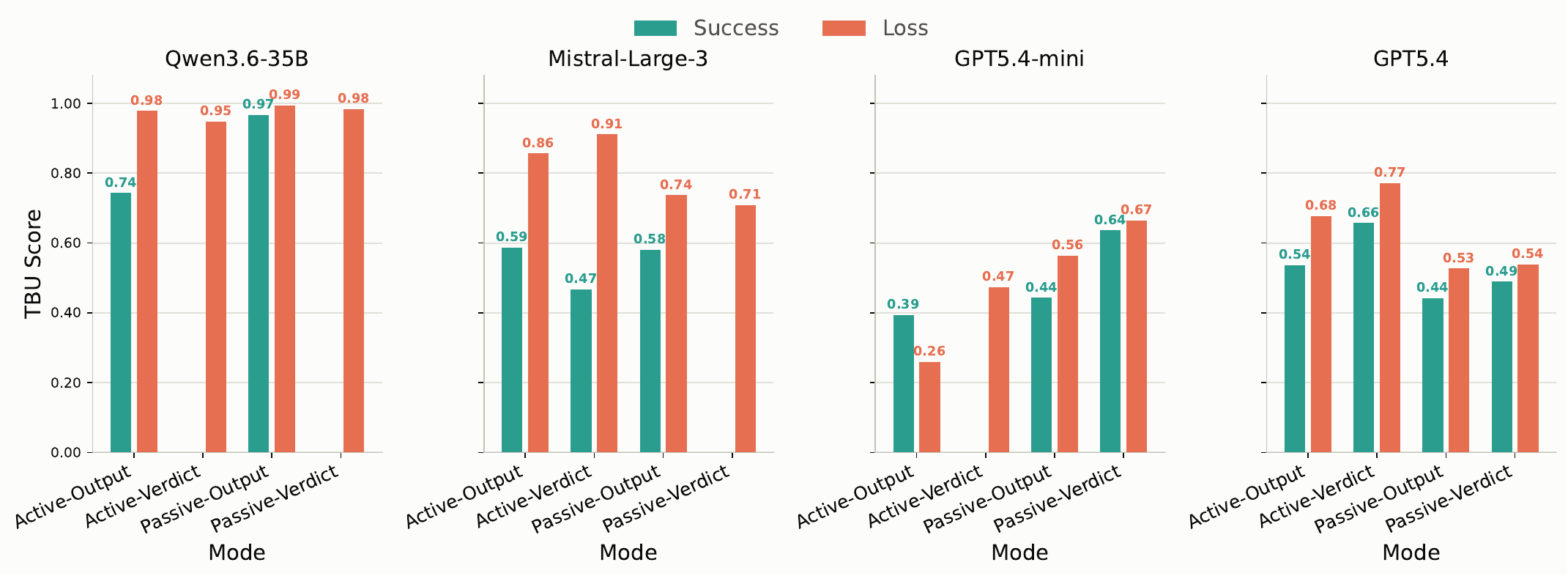}
  \caption{TBU scores across models and interaction modes. Higher scores indicate more of the turn budget was used; scores below 0.67 correspond to fewer than 10 turns. Missing bars indicate no instances for that outcome.
  }
  \label{fig:tbuscore}
  \vspace{-0.3cm}  
\end{figure*}

\section{Results}
\subsection{Analysing model capabilities}
\label{subsec:modelsuccessanalysis}
\textit{Can LLMs reconstruct a hidden Python function from input-output evidence?} Figure~\ref{fig:overallsuccessrate} presents the success rates across the evaluated models and interaction modes~\footnote{Success rates across domains are available in Figures~\ref{fig:successrate-1} and~\ref{fig:successrate-3} in the Appendix~\ref{sec:appendix-quantanalysis}.}. The overall success rates range from $0.03$ for Qwen3.6-35B, through $0.20$ for Mistral-Large-3, to $0.64$ for GPT5.4. The low success rates of Qwen3.6-35B and Mistral-Large-3 are mainly because both models continue probing until the final turn, exhausting all $15$ turns without converging to a solution. Moreover, Qwen3.6-35B struggles to follow the intended response format, with parser errors occurring in $40.64\%$ of all instances (see Table~\ref{tab:parsererrors} in the Appendix). Together, these results suggest that failures arise from difficulties with induction and format adherence.

We then analyse how each interaction mode contributes to the model-overall success rates. We use the single-turn modes as baselines for comparison with the corresponding multi-turn modes. For most models, across both feedback types, success rates follow the order single-turn, passive, and active, indicating that models perform better when evidence is provided upfront than when they must acquire it through interaction. Between the two single-turn modes, the single-turn-output mode achieves higher success rates. Membership feedback is more challenging because it only indicates whether a proposed output is correct, whereas output feedback reveals the corresponding output and is the direct evidence about the hidden function. This challenge is greater in the active-verdict mode, where the model must also select the queries.

Overall, the results show that the models are capable of performing the task, but their performance remains far from saturated.

\begin{figure*}[t]
\centering
  \vspace{-0.2cm} 
  \includegraphics[width=0.88\textwidth]{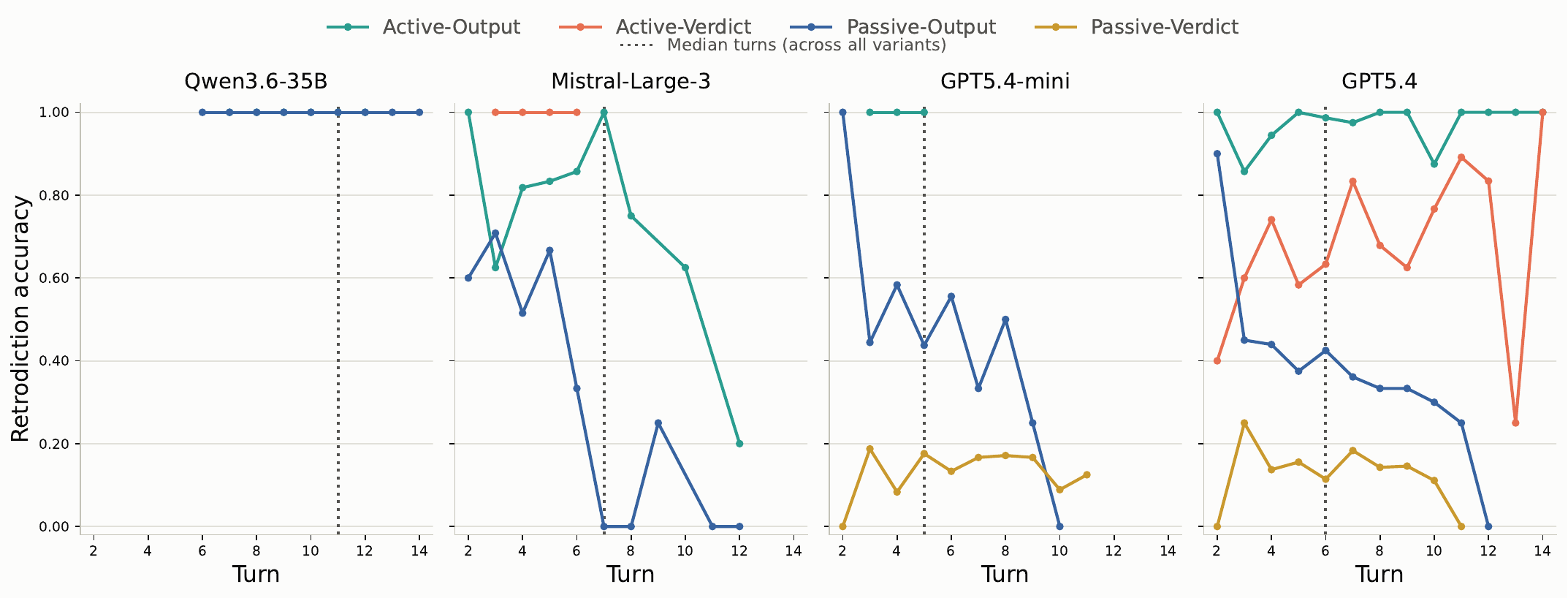}
  \caption{Hypothesis retrodiction accuracy across turns. Dotted lines show the median turn count for each model.
  }
  \label{fig:retrodiction_allmodes}
  \vspace{-0.3cm}  
\end{figure*}

\begin{figure*}[t]
\centering
  \includegraphics[width=0.88\textwidth]{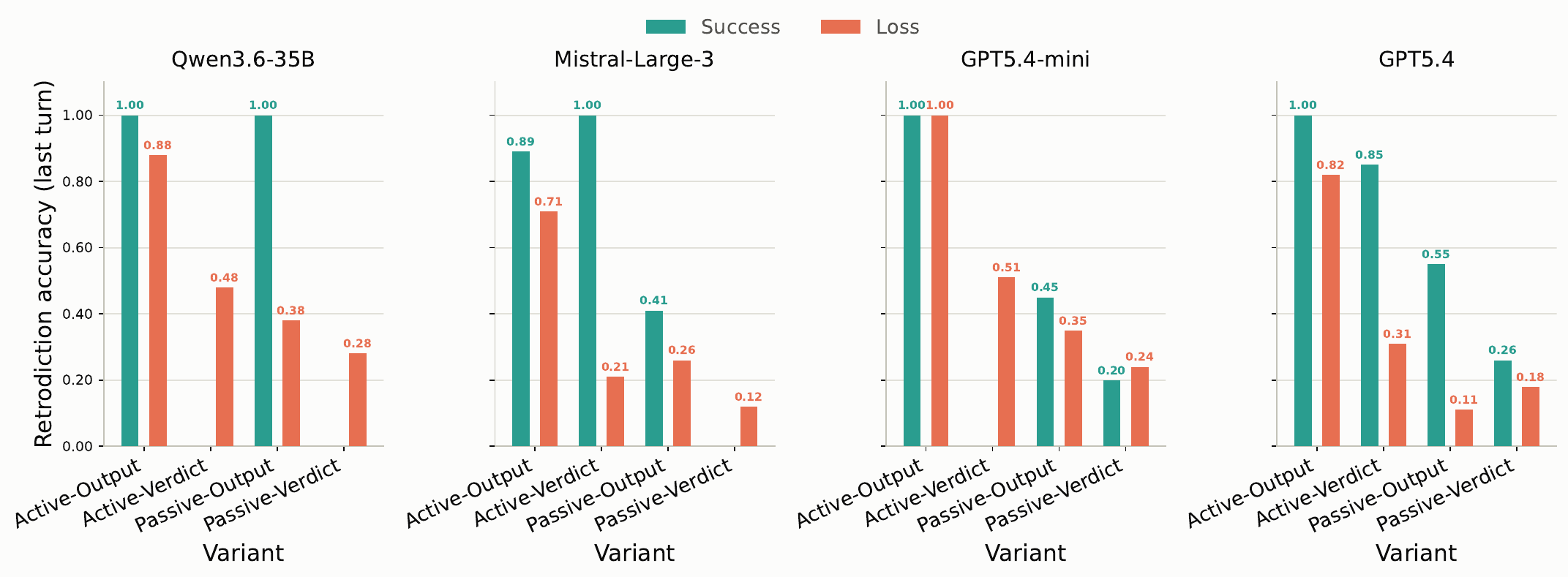}
  \caption{Hypothesis retrodiction accuracy at the final turn across models and interaction modes.
  }
  \label{fig:retrodiction_lastturn}
  \vspace{-0.3cm}  
\end{figure*}

\subsection{Evidence-budget management}
\label{subsec:evidence-budget}
\textit{How does distributing evidence across turns, with model-controlled stopping, affect success and use of the available turn budget?}
Figure~\ref{fig:tbuscore} presents the TBU scores for each model and interaction mode. Qwen3.6-35B, Mistral-Large-3, and GPT5.4-mini have a score of zero in at least one interaction mode. Across most of the models, successful instances have lower TBU scores than failed instances. In active-output and passive-output, GPT5.4-mini has lower TBU scores than the other models, at $0.39$ and $0.44$, respectively. Similarly, Mistral-Large-3 has a lower TBU score than the other models in active-verdict, at $0.47$. However, TBU measures how much of the available turn budgets used before an episode ends and should therefore be considered together with the overall success rates.

The higher TBU scores for failed instances suggest that these instances are more challenging and lead to longer interactions. Qwen3.6-35B and to an extent Mistral-Large-3 nearly exhausts the turn budget, whereas the other models use fewer than 10 turns in most interaction modes, as evident from TBU scores below $0.67$.

This pattern shows that models often make their guesses before using the full turn budget. However, earlier commitment does not translate into higher success rates, as all models achieve lower success rates. To examine whether this difference is related to model-controlled stopping, we compare single-turn-output and passive-output. GPT5.4 and GPT5.4-mini achieve similar success rates in single-turn-output, at $0.82$ and $0.74$, respectively, but differ in passive-output, at $0.78$ and $0.40$ (Figure~\ref{fig:overallsuccessrate}). The larger gap in passive-output suggests that model-controlled evidence acquisition and stopping affect the two models differently. Overall, some models stop before using the available evidence and fail, whereas others exhaust the turn budget without converging to a solution.

\paragraph{Hypothesis Retrodiction Accuracy} Figure~\ref{fig:retrodiction_allmodes} shows how hypothesis retrodiction accuracy scores vary across settings. In active-output, most models begin with high scores, decline over the intermediate turns, and then increase again toward the final turns. In the other modes, the scores generally decrease as the interaction progresses. This suggests that hypotheses are matched more accurately during the initial turns and weakens as additional evidence are available. To examine whether this pattern persists at the point of commitment, Figure~\ref{fig:retrodiction_lastturn} reports the HRA at the final turn. For the successful instances of active interaction modes, the final-turn HRA ranges from $0.85$ to $1.00$ across models, whereas for the other modes it ranges from $0.20$ to $0.55$. Overall, this suggests that models may struggle to validate their hypotheses against the accumulated evidence and may commit before observing enough evidence to refine them. We next examine whether this pattern depends on how evidence is acquired.

\begin{figure*}[t]
\centering
  \vspace{-0.3cm}  
  \includegraphics[width=0.88\textwidth]{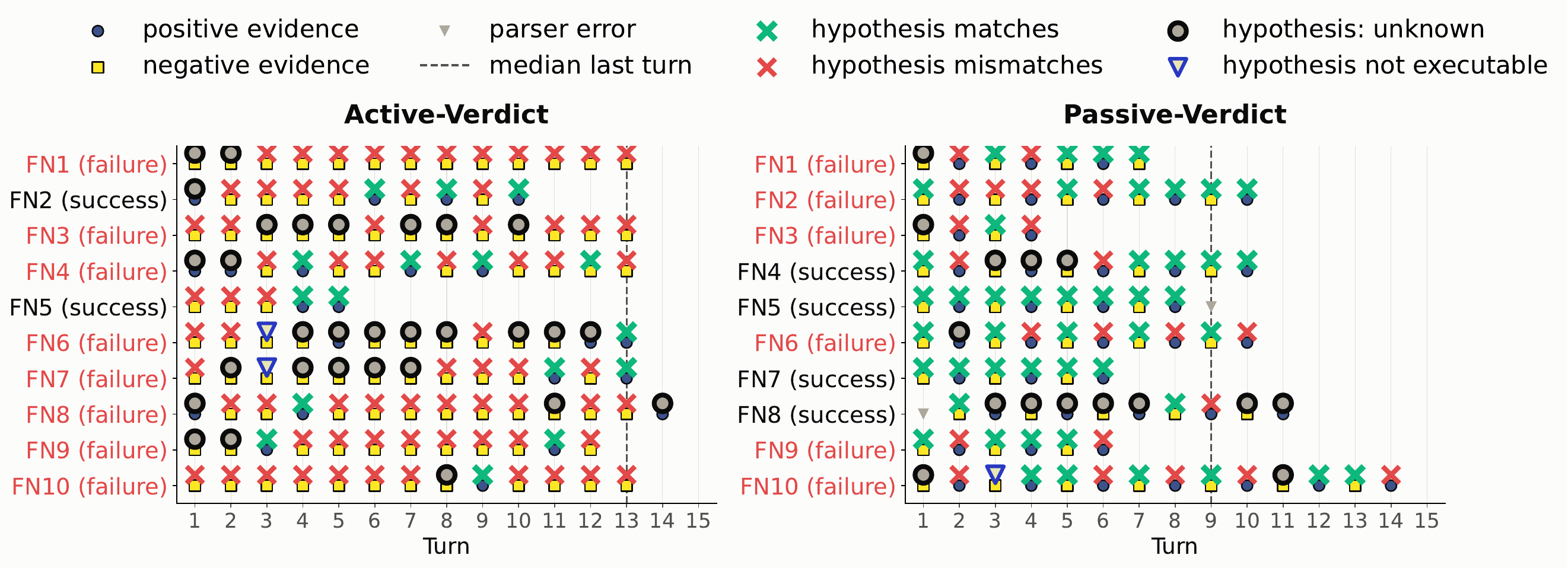}
  \caption{Negative evidence impact on task success for GPT5.4 in Numbers domain.
  }
  \label{fig:qualitativeanalysis-fig3}
\end{figure*}

\subsection{Query selection}
\label{subsec:queryselection}
\textit{Does self-selection affect stopping behaviour and overall success?} We examine this by comparing the active-output and passive-output modes. Both are multi-turn, but active-output requires the model to select each query, whereas in passive-output the Game Master provides the next example. As shown in Figure~\ref{fig:tbuscore}, Qwen3.6-35B, Mistral-Large-3, and GPT5.4-mini have higher TBU scores in active-output than in passive-output for successful instances, while GPT5.4 shows the opposite pattern. For failed instances, only GPT5.4-mini has a lower TBU score in active-output. Despite these differences, most of the models achieve higher success rates in passive-output. At the final turn, hypothesis retrodiction accuracy (see Figure~\ref{fig:retrodiction_lastturn}) is higher in active-output for all models, but this accuracy is measured against examples selected by the model and may therefore reflect consistency with its current hypothesis rather than the ability of those examples to distinguish it from competing hypotheses. In passive-output, lower retrodiction accuracy and higher TBU scores suggest that models observe more Game Master-provided examples that challenge their current hypotheses before committing. Overall, self-selection may lead models to stop earlier on examples that fit their current hypotheses, while passive presentation exposes them to more evidence and leads to higher success.

\begin{figure}[t]
\centering
  \vspace{-0.3cm}  
  \includegraphics[width=0.49\textwidth]{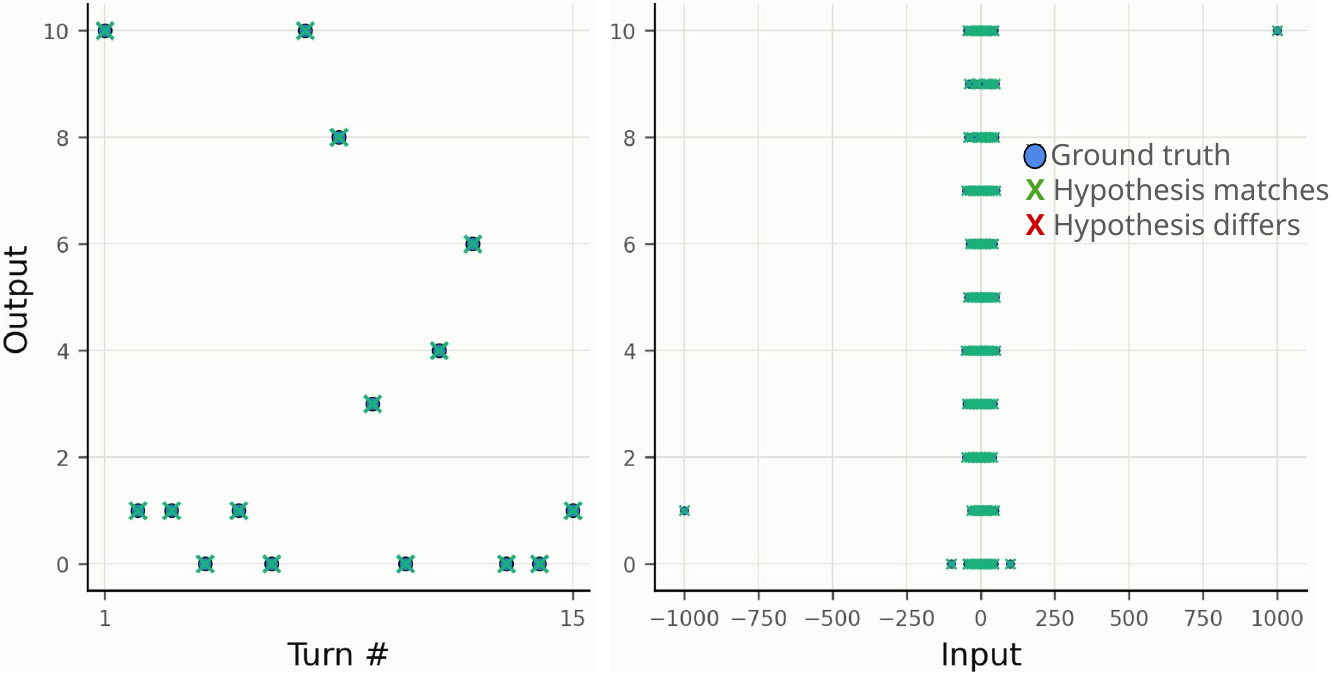}
  \caption{7.56\% of failed instances of GPT5.4-mini's end with a correct hypothesis but no submitted solution.
  }
  \label{fig:qualitativeanalysis-fig1}
  \vspace{-0.3cm}  
\end{figure}

\section{Qualitative Analysis}
\label{sec:qualitativeanalysis}
\begin{figure}[t]
\centering
  \vspace{-0.3cm}  
  \includegraphics[width=0.49\textwidth]{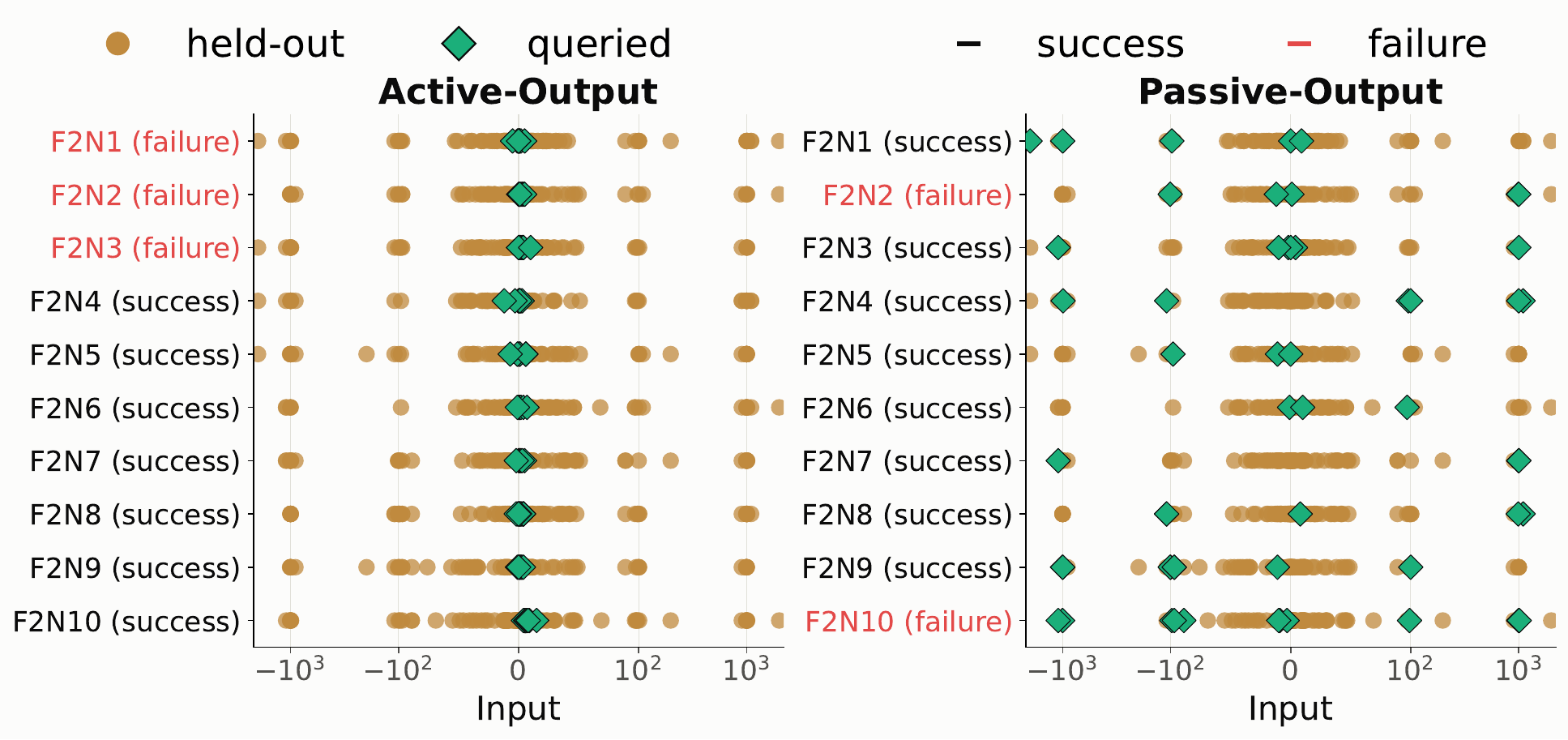}
  \caption{\textbf{Spread of input queries:} These results are for GPT5.4 model, for the two\_numbers category. The narrower coverage of Active-output queries may contribute to the higher failure rate.
  }
  \label{fig:qualitativeanalysis-fig2}
  \vspace{-0.3cm}  
\end{figure}

We analyse failed interactions to identify patterns across models and modes. Figure~\ref{fig:qualitativeanalysis-fig3} shows how negative evidence affects task success for GPT5.4. Active-verdict requires more turns than passive-verdict, consistent with the higher TBU score of the latter discussed in Section~\ref{subsec:evidence-budget}. Moreover, in active-verdict, when the Game Master labels an input-output pair as \textsc{False}, the model changes both its subsequent queries and hypotheses, suggesting that it eliminates some possibilities. However, these revisions do not consistently lead to a hypothesis that fits the accumulated evidence, which may explain the longer interactions and lower success. As shown in Figure~\ref{fig:qualitativeanalysis-fig1}, some failures are \textit{unclaimed wins}: the model reaches a correct hypothesis but continues querying until the turn budget is exhausted~\footnote{Details across models are in Table~\ref{tab:failstats} in the Appendix}. This suggests that some models may not reliably assess when their hypotheses are sufficiently supported, leading either to premature commitment or failure to converge. Figure~\ref{fig:qualitativeanalysis-fig2} compares the input coverage when GPT5.4 selects queries with that of examples provided by the Game Master. The active queries cover a narrower region, whereas the Game Master-provided examples span a wider range. This supports the analysis in Section~\ref{subsec:queryselection} that self-selection can limit input coverage and reduce success in active-output.  Together, these examples suggest that task success depends on query coverage, hypothesis validation, and how models use negative evidence. Additional turn-level traces showing the selected actions and corresponding outputs for each model, along with analyses of confidence scores, interaction states, and query rationales, are available in the Appendix~\ref{sec:appendix-qualitative-analysis}. 

\section{Conclusion}
We introduce Alien Abduction, a multi-turn dialogue game for exploring the abductive reasoning capabilities of LLMs across interaction settings and models. Our findings show that the main difficulty lies in determining how much evidence to collect and how to use it to revise hypotheses over time. Models often commit before gathering enough evidence or continue querying without converging. Therefore the probe reveals behaviours that are not visible in single-turn evaluations and provides a way to examine how models manage evidence before reaching a final answer. Overall, the results suggest that improving abductive reasoning requires better query selection, hypothesis validation, and stopping.

\section*{Limitations}
Our study makes several design choices to support controlled comparisons across interaction modes. These choices involve deliberate trade-offs: they strengthen experimental control while limiting the range of settings examined.

\textbf{1. Task scope.} Alien Abduction probe uses synthetic functions across five controlled domains. This design isolates evidence acquisition, hypothesis revision, and stopping behaviour. However, open-ended abductive reasoning may involve ambiguity, noisy evidence, and background knowledge not represented in the current tasks. Our findings therefore concern reasoning over controlled rule-induction tasks, and future work can examine whether the observed behaviours extend to real-world settings.

\textbf{2. Target generation and model coverage.} Candidate target functions are generated using GPT5.4, automatically validated, and randomly sampled from the resulting pool. GPT5.4 contributes only the initial candidates and does not select the final targets or evaluate model responses. All target functions are checked to ensure that they run without errors and produce deterministic outputs, and every model is evaluated on the same targets and test cases. These controls reduce the possibility of an advantage for GPT5.4. However, the resulting function distribution may still reflect patterns more commonly produced by that model. In addition, the evaluated models do not cover the full range of available LLMs. Future work could extend the framework by using targets produced by multiple generators and evaluating additional models and prompting settings.

\textbf{3. Interaction settings.} The interactive modes use a fixed turn budget, and each mode specifies who selects the examples and what information is returned. Keeping these settings constant enables direct comparisons between active, passive, and single-turn interaction. Different turn budgets or Game Master selection policies nevertheless affect query selection, stopping behaviour, and task success. Examining these factors would provide a broader account of turn-budget management.

\textbf{4. Behavioural traces.} HRA is computed from the hypotheses explicitly reported by the models. These reports provide a turn-level view of whether the stated hypothesis remains consistent with the accumulated evidence, but they may not fully represent the model's internal belief state. Confidence scores, interaction states, and query rationales should similarly be interpreted as observable behavioural traces rather than faithful explanations of the underlying reasoning process.

% \bibliographystyle is already set by acl.sty
\bibliography{custom}

@inproceedings{honovich2023instruction,
  author    = {Honovich, Or and Shaham, Uri and Bowman, Samuel R. and Levy, Omer},
  title     = {Instruction Induction: From Few Examples to Natural Language Task Descriptions},
  booktitle = {Proceedings of the 61st Annual Meeting of the Association for
               Computational Linguistics (Volume 1: Long Papers)},
  month     = jul,
  year      = {2023},
  address   = {Toronto, Canada},
  publisher = {Association for Computational Linguistics},
  pages     = {1935--1952},
  doi       = {10.18653/v1/2023.acl-long.108},
  url       = {https://aclanthology.org/2023.acl-long.108/}
}

@inproceedings{sun2024itd,
  author    = {Sun, Wangtao and Xu, Haotian and Yu, Xuanqing and Chen, Pei and
               He, Shizhu and Zhao, Jun and Liu, Kang},
  title     = {{ItD}: Large Language Models Can Teach Themselves Induction
               through Deduction},
  booktitle = {Proceedings of the 62nd Annual Meeting of the Association for
               Computational Linguistics (Volume 1: Long Papers)},
  month     = aug,
  year      = {2024},
  address   = {Bangkok, Thailand},
  publisher = {Association for Computational Linguistics},
  pages     = {2719--2731},
  doi       = {10.18653/v1/2024.acl-long.150},
  url       = {https://aclanthology.org/2024.acl-long.150/}
}

@inproceedings{he2025idea,
  author    = {He, Kaiyu and Zhang, Mian and Yan, Shuo and Wu, Peilin and
               Chen, Zhiyu Zoey},
  title     = {{IDEA}: Enhancing the Rule Learning Ability of Large Language
               Model Agent through Induction, Deduction, and Abduction},
  booktitle = {Findings of the Association for Computational Linguistics:
               {ACL} 2025},
  month     = jul,
  year      = {2025},
  address   = {Vienna, Austria},
  publisher = {Association for Computational Linguistics},
  pages     = {13563--13597},
  doi       = {10.18653/v1/2025.findings-acl.698},
  url       = {https://aclanthology.org/2025.findings-acl.698/}
}

@inproceedings{cifka2023context,
  author    = {C{\'i}fka, Ond{\v{r}}ej and Liutkus, Antoine},
  title     = {Black-Box Language Model Explanation by Context Length Probing},
  booktitle = {Proceedings of the 61st Annual Meeting of the Association for
               Computational Linguistics (Volume 2: Short Papers)},
  month     = jul,
  year      = {2023},
  address   = {Toronto, Canada},
  publisher = {Association for Computational Linguistics},
  pages     = {1067--1079},
  doi       = {10.18653/v1/2023.acl-short.92},
  url       = {https://aclanthology.org/2023.acl-short.92/}
}

@inproceedings{yin2026oracle,
  author    = {Yin, Congchi and Wu, Tianyi and Shu, Yankai and Gu, Alex and
               Wang, Yunhan and Shao, Jun and Jiang, Xun and Li, Piji},
  title     = {Investigating Advanced Reasoning of Large Language Models via
               Black-Box Environment Interaction},
  booktitle = {Proceedings of the 43rd International Conference on Machine Learning},
  series    = {Proceedings of Machine Learning Research},
  volume    = {306},
  publisher = {PMLR},
  year      = {2026}
}

@inproceedings{chalamalasetti-etal-2023-clembench,
    title = "clembench: Using Game Play to Evaluate Chat-Optimized Language Models as Conversational Agents",
    author = {Chalamalasetti, Kranti  and
      G{\"o}tze, Jana  and
      Hakimov, Sherzod  and
      Madureira, Brielen  and
      Sadler, Philipp  and
      Schlangen, David},
    editor = "Bouamor, Houda  and
      Pino, Juan  and
      Bali, Kalika",
    booktitle = "Proceedings of the 2023 Conference on Empirical Methods in Natural Language Processing",
    month = dec,
    year = "2023",
    address = "Singapore",
    publisher = "Association for Computational Linguistics",
    url = "https://aclanthology.org/2023.emnlp-main.689/",
    doi = "10.18653/v1/2023.emnlp-main.689",
    pages = "11174--11219"
}

@article{chari2026wild,
  author  = {Chari, Anirudh and Pattanaik, Neil},
  title   = {Wild Guesses and Mild Guesses in Active Concept Learning},
  journal = {CoRR},
  volume  = {abs/2602.06818},
  year    = {2026},
  doi     = {10.48550/arXiv.2602.06818},
  url     = {https://arxiv.org/abs/2602.06818}
}

@article{chen2021evaluating,
  author  = {Chen, Mark and others},
  title   = {Evaluating Large Language Models Trained on Code},
  journal = {CoRR},
  volume  = {abs/2107.03374},
  year    = {2021},
  doi     = {10.48550/arXiv.2107.03374},
  url     = {https://arxiv.org/abs/2107.03374}
}

@article{austin2021program,
  author  = {Austin, Jacob and Odena, Augustus and Nye, Maxwell I. and
             Bosma, Maarten and Michalewski, Henryk and Dohan, David and
             Jiang, Ellen and Cai, Carrie J. and Terry, Michael and
             Le, Quoc V. and Sutton, Charles},
  title   = {Program Synthesis with Large Language Models},
  journal = {CoRR},
  volume  = {abs/2108.07732},
  year    = {2021},
  doi     = {10.48550/arXiv.2108.07732},
  url     = {https://arxiv.org/abs/2108.07732}
}

@inproceedings{nijkamp2023codegen2,
  author    = {Nijkamp, Erik and Hayashi, Hiroaki and Xiong, Caiming and
               Savarese, Silvio and Zhou, Yingbo},
  title     = {{CodeGen2}: Lessons for Training {LLM}s on Programming and
               Natural Languages},
  booktitle = {The Eleventh International Conference on Learning Representations},
  year      = {2023}
}

@inproceedings{DBLP:conf/nips/YanLLYFSDC25,
  author       = {Kai Yan and
                  Zhan Ling and
                  Kang Liu and
                  Yifan Yang and
                  Ting{-}Han Fan and
                  Lingfeng Shen and
                  Zhengyin Du and
                  Jiecao Chen},
  editor       = {Danielle Belgrave and
                  Cheng Zhang and
                  Laura N. Montoya and
                  Hsuan{-}Tien Lin and
                  Razvan Pascanu and
                  Piotr Koniusz and
                  Marzyeh Ghassemi and
                  Nancy Chen and
                  Iv{\'{a}}n Vladimir Meza Ru{\'{\i}}z and
                  Arturo Loaiza{-}Bonilla},
  title        = {MIR-Bench: Can Your {LLM} Recognize Complicated Patterns via Many-Shot
                  In-Context Reasoning?},
  booktitle    = {Advances in Neural Information Processing Systems 38: Annual Conference
                  on Neural Information Processing Systems 2025, NeurIPS 2025, San Diego,
                  CA, USA, December 2-7, 2025 / Mexico City, Mexico, November 30 - December
                  5, 2025},
  year         = {2025},
  url          = {http://papers.nips.cc/paper\_files/paper/2025/hash/796076672b00f54fb01d05a2e5fde363-Abstract-Datasets\_and\_Benchmarks\_Track.html},
  bibsource    = {dblp computer science bibliography, https://dblp.org}
}

@inproceedings{DBLP:conf/acl/XuC00LHHC025,
  author       = {Ruiyang Xu and
                  Jialun Cao and
                  Yaojie Lu and
                  Ming Wen and
                  Hongyu Lin and
                  Xianpei Han and
                  Ben He and
                  Shing{-}Chi Cheung and
                  Le Sun},
  editor       = {Wanxiang Che and
                  Joyce Nabende and
                  Ekaterina Shutova and
                  Mohammad Taher Pilehvar},
  title        = {{CRUXEVAL-X:} {A} Benchmark for Multilingual Code Reasoning, Understanding
                  and Execution},
  booktitle    = {Proceedings of the 63rd Annual Meeting of the Association for Computational
                  Linguistics (Volume 1: Long Papers), {ACL} 2025, Vienna, Austria,
                  July 27 - August 1, 2025},
  pages        = {23762--23779},
  publisher    = {Association for Computational Linguistics},
  year         = {2025},
  url          = {https://doi.org/10.18653/v1/2025.acl-long.1158},
  doi          = {10.18653/V1/2025.ACL-LONG.1158},
  bibsource    = {dblp computer science bibliography, https://dblp.org}
}

@article{zheng2023survey,
  author  = {Zheng, Zibin and Ning, Kaiwen and Wang, Yanlin and
             Zhang, Jingwen and Zheng, Dewu and Ye, Mingxi and Chen, Jiachi},
  title   = {A Survey of Large Language Models for Code: Evolution,
             Benchmarking, and Future Trends},
  journal = {CoRR},
  volume  = {abs/2311.10372},
  year    = {2023},
  doi     = {10.48550/arXiv.2311.10372},
  url     = {https://arxiv.org/abs/2311.10372}
}

@inproceedings{gu2024cruxeval,
  author    = {Gu, Alex and Roziere, Baptiste and Leather, Hugh James and
               Solar-Lezama, Armando and Synnaeve, Gabriel and Wang, Sida},
  title     = {{CRUXEval}: A Benchmark for Code Reasoning, Understanding and
               Execution},
  booktitle = {Proceedings of the 41st International Conference on Machine Learning},
  pages     = {16568--16621},
  year      = {2024},
  volume    = {235},
  series    = {Proceedings of Machine Learning Research},
  publisher = {PMLR},
  url       = {https://proceedings.mlr.press/v235/gu24c.html}
}

@inproceedings{wei2025codearc,
  author    = {Wei, Anjiang and Suresh, Tarun and Cao, Jiannan and
               Kannan, Naveen and Wu, Yuheng and Yan, Kai and
               Teixeira, Thiago S. F. X. and Wang, Ke and Aiken, Alex},
  title     = {{CodeARC}: Benchmarking Reasoning Capabilities of {LLM} Agents
               for Inductive Program Synthesis},
  booktitle = {Second Conference on Language Modeling},
  year      = {2025},
  url       = {https://openreview.net/forum?id=Q5pVZCrrKr}
}

@inproceedings{lee2025syntra,
  author    = {Lee, Kang-il and Koo, Jahyun and Yoon, Seunghyun and
               Kim, Minbeom and Koh, Hyukhun and Lee, Dongryeol and
               Jung, Kyomin},
  title     = {Program Synthesis via Test-Time Transduction},
  booktitle = {Advances in Neural Information Processing Systems},
  volume    = {38},
  year      = {2025},
  publisher = {Curran Associates, Inc.}
}

@inproceedings{surana2026iprog,
  author    = {Surana, Shraddha and Srinivasan, Ashwin and Bain, Michael},
  title     = {Engineering Systems for Data Analysis Using Interactive Structured
               Inductive Programming},
  booktitle = {Advanced Information Systems Engineering: 38th International
               Conference, {CAiSE} 2026, Verona, Italy, June 8--12, 2026,
               Proceedings, Part {I}},
  series    = {Lecture Notes in Computer Science},
  volume    = {16558},
  pages     = {249--266},
  publisher = {Springer},
  year      = {2026},
  doi       = {10.1007/978-3-032-28110-4_14}
}

@article{geng2025reliable,
  author  = {Geng, Jiayi and Chen, Howard and Arumugam, Dilip and
             Griffiths, Thomas L.},
  title   = {Are Large Language Models Reliable {AI} Scientists?
             Assessing Reverse-Engineering of Black-Box Systems},
  journal = {CoRR},
  volume  = {abs/2505.17968},
  year    = {2025},
  doi     = {10.48550/arXiv.2505.17968},
  url     = {https://arxiv.org/abs/2505.17968}
}

@inproceedings{zhou2025arbench,
  author    = {Zhou, Zhanke and Feng, Xiao and Zhu, Zhaocheng and
               Yao, Jiangchao and Koyejo, Sanmi and Han, Bo},
  title     = {From Passive to Active Reasoning: Can Large Language Models Ask
               the Right Questions under Incomplete Information?},
  booktitle = {Proceedings of the 42nd International Conference on Machine Learning},
  volume    = {267},
  series    = {Proceedings of Machine Learning Research},
  publisher = {PMLR},
  year      = {2025},
  url       = {https://proceedings.mlr.press/v267/zhou25e.html}
}

@inproceedings{chen2025physgym,
  author    = {Chen, Yimeng and Piekos, Piotr and Ostaszewski, Mateusz and
               Laakom, Firas and Schmidhuber, J{\"u}rgen},
  title     = {{PHYSGYM}: Benchmarking {LLM}s in Interactive Physics Discovery
               with Controlled Priors},
  booktitle = {Advances in Neural Information Processing Systems},
  volume    = {38},
  year      = {2025},
  publisher = {Curran Associates, Inc.}
}

@article{singh2023explaining,
  author  = {Singh, Chandan and Hsu, Aliyah R. and Antonello, Richard J. and
             Jain, Shailee and Huth, Alexander G. and Yu, Bin and Gao, Jianfeng},
  title   = {Explaining Black Box Text Modules in Natural Language with
             Language Models},
  journal = {CoRR},
  volume  = {abs/2305.09863},
  year    = {2023},
  doi     = {10.48550/arXiv.2305.09863},
  url     = {https://arxiv.org/abs/2305.09863}
}

@book{peirce1934collected,
  title={Collected papers of charles sanders peirce},
  author={Peirce, Charles Sanders},
  volume={5},
  year={1934},
  publisher={Harvard University Press}
}

@misc{qwen36_35b_a3b,
    title = {{Qwen3.6-35B-A3B}: Agentic Coding Power, Now Open to All},
    url = {https://qwen.ai/blog?id=qwen3.6-35b-a3b},
    author = {{Qwen Team}},
    month = {April},
    year = {2026}
}

@article{DBLP:journals/corr/abs-2601-08584,
  author       = {Mistral AI},
  title        = {Ministral 3},
  journal      = {CoRR},
  volume       = {abs/2601.08584},
  year         = {2026},
  url          = {https://doi.org/10.48550/arXiv.2601.08584},
  doi          = {10.48550/ARXIV.2601.08584},
  eprinttype   = {arXiv},
  eprint       = {2601.08584},
  bibsource    = {dblp computer science bibliography, https://dblp.org}
}

@inproceedings{balloccu-etal-2024-leak,
    title = "Leak, Cheat, Repeat: Data Contamination and Evaluation Malpractices in Closed-Source {LLM}s",
    author = "Balloccu, Simone  and
      Schmidtov{\'a}, Patr{\'i}cia  and
      Lango, Mateusz  and
      Dusek, Ondrej",
    editor = "Graham, Yvette  and
      Purver, Matthew",
    booktitle = "Proceedings of the 18th Conference of the European Chapter of the Association for Computational Linguistics (Volume 1: Long Papers)",
    month = mar,
    year = "2024",
    address = "St. Julian{'}s, Malta",
    publisher = "Association for Computational Linguistics",
    url = "https://aclanthology.org/2024.eacl-long.5/",
    doi = "10.18653/v1/2024.eacl-long.5",
    pages = "67--93"
}

@inproceedings{ICLR2024_46e624c2,
 author = {Oren, Yonatan and Meister, Nicole and Chatterji, Niladri and Ladhak, Faisal and Hashimoto, Tatsunori},
 booktitle = {International Conference on Learning Representations},
 editor = {B. Kim and Y. Yue and S. Chaudhuri and K. Fragkiadaki and M. Khan and Y. Sun},
 pages = {16354--16372},
 title = {Proving Test Set Contamination in Black-Box Language Models},
 url = {https://proceedings.iclr.cc/paper_files/paper/2024/file/46e624c244cff669223d488defd4e835-Paper-Conference.pdf},
 volume = {2024},
 year = {2024}
}

@article{DBLP:journals/corr/abs-2502-14425,
  author       = {Yuxing Cheng and
                  Yi Chang and
                  Yuan Wu},
  title        = {A Survey on Data Contamination for Large Language Models},
  journal      = {CoRR},
  volume       = {abs/2502.14425},
  year         = {2025},
  url          = {https://doi.org/10.48550/arXiv.2502.14425},
  doi          = {10.48550/ARXIV.2502.14425},
  eprinttype   = {arXiv},
  eprint       = {2502.14425},
  bibsource    = {dblp computer science bibliography, https://dblp.org}
}

@inproceedings{NEURIPS2022_18abbeef,
 author = {Lewkowycz, Aitor and Andreassen, Anders and Dohan, David and Dyer, Ethan and Michalewski, Henryk and Ramasesh, Vinay and Slone, Ambrose and Anil, Cem and Schlag, Imanol and Gutman-Solo, Theo and Wu, Yuhuai and Neyshabur, Behnam and Gur-Ari, Guy and Misra, Vedant},
 booktitle = {Advances in Neural Information Processing Systems},
 doi = {10.52202/068431-0278},
 editor = {S. Koyejo and S. Mohamed and A. Agarwal and D. Belgrave and K. Cho and A. Oh},
 pages = {3843--3857},
 publisher = {Curran Associates, Inc.},
 title = {Solving Quantitative Reasoning Problems with Language Models},
 url = {https://proceedings.neurips.cc/paper_files/paper/2022/file/18abbeef8cfe9203fdf9053c9c4fe191-Paper-Conference.pdf},
 volume = {35},
 year = {2022}
}

@article{DBLP:journals/corr/abs-2407-21783,
  author       = {Llama Team},
  title        = {The Llama 3 Herd of Models},
  journal      = {CoRR},
  volume       = {abs/2407.21783},
  year         = {2024},
  url          = {https://doi.org/10.48550/arXiv.2407.21783},
  doi          = {10.48550/ARXIV.2407.21783},
  eprinttype   = {arXiv},
  eprint       = {2407.21783},
  bibsource    = {dblp computer science bibliography, https://dblp.org}
}

@inproceedings{NEURIPS2024_53384f20,
 author = {Zhang, Hugh and Da, Jeff and Lee, Dean and Robinson, Vaughn and Wu, Catherine and Song, Will and Zhao, Tiffany and Raja, Pranav and Zhuang, Charlotte and Slack, Dylan and Lyu, Qin and Hendryx, Sean and Kaplan, Russell and Lunati, Michele and Yue, Summer},
 booktitle = {Advances in Neural Information Processing Systems},
 doi = {10.52202/079017-1485},
 editor = {A. Globerson and L. Mackey and D. Belgrave and A. Fan and U. Paquet and J. Tomczak and C. Zhang},
 pages = {46819--46836},
 publisher = {Curran Associates, Inc.},
 title = {A Careful Examination of Large Language Model Performance on Grade School Arithmetic},
 url = {https://proceedings.neurips.cc/paper_files/paper/2024/file/53384f2090c6a5cac952c598fd67992f-Paper-Datasets_and_Benchmarks_Track.pdf},
 volume = {37},
 year = {2024}
}

@article{wason1960failure,
  title={On the failure to eliminate hypotheses in a conceptual task},
  author={Wason, Peter C},
  journal={Quarterly journal of experimental psychology},
  volume={12},
  number={3},
  pages={129--140},
  year={1960},
  publisher={SAGE Publications Sage UK: London, England}
}

@book{bruner2017study,
  title={A study of thinking},
  author={Bruner, Jerome},
  year={2017},
  publisher={Routledge}
}

@inproceedings{DBLP:conf/nips/BrownMRSKDNSSAA20,
  author       = {Tom B. Brown and
                  Benjamin Mann and
                  Nick Ryder and
                  Melanie Subbiah and
                  Jared Kaplan and
                  et al
},
  editor       = {Hugo Larochelle and
                  Marc'Aurelio Ranzato and
                  Raia Hadsell and
                  Maria{-}Florina Balcan and
                  Hsuan{-}Tien Lin},
  title        = {Language Models are Few-Shot Learners},
  booktitle    = {Advances in Neural Information Processing Systems 33: Annual Conference
                  on Neural Information Processing Systems 2020, NeurIPS 2020, December
                  6-12, 2020, virtual},
  year         = {2020},
  url          = {https://proceedings.neurips.cc/paper/2020/hash/1457c0d6bfcb4967418bfb8ac142f64a-Abstract.html},
  bibsource    = {dblp computer science bibliography, https://dblp.org}
}

@inproceedings{DBLP:conf/nips/Wei0SBIXCLZ22,
  author       = {Jason Wei and
                  Xuezhi Wang and
                  Dale Schuurmans and
                  Maarten Bosma and
                  Brian Ichter and
                  Fei Xia and
                  Ed H. Chi and
                  Quoc V. Le and
                  Denny Zhou},
  editor       = {Sanmi Koyejo and
                  S. Mohamed and
                  A. Agarwal and
                  Danielle Belgrave and
                  K. Cho and
                  A. Oh},
  title        = {Chain-of-Thought Prompting Elicits Reasoning in Large Language Models},
  booktitle    = {Advances in Neural Information Processing Systems 35: Annual Conference
                  on Neural Information Processing Systems 2022, NeurIPS 2022, New Orleans,
                  LA, USA, November 28 - December 9, 2022},
  year         = {2022},
  url          = {http://papers.nips.cc/paper\_files/paper/2022/hash/9d5609613524ecf4f15af0f7b31abca4-Abstract-Conference.html},
  bibsource    = {dblp computer science bibliography, https://dblp.org}
}

@article{DBLP:journals/csur/LiuYFJHN23,
  author       = {Pengfei Liu and
                  Weizhe Yuan and
                  Jinlan Fu and
                  Zhengbao Jiang and
                  Hiroaki Hayashi and
                  Graham Neubig},
  title        = {Pre-train, Prompt, and Predict: {A} Systematic Survey of Prompting
                  Methods in Natural Language Processing},
  journal      = {{ACM} Comput. Surv.},
  volume       = {55},
  number       = {9},
  pages        = {195:1--195:35},
  year         = {2023},
  url          = {https://doi.org/10.1145/3560815},
  doi          = {10.1145/3560815},
  bibsource    = {dblp computer science bibliography, https://dblp.org}
}

\FloatBarrier
\appendix
\section{Benchmark Functions}
\label{app:functions}
Table~\ref{tab:alien_abduction_functions} shows the 50 target functions used for evaluation. The functions are grouped by domain, and those within each domain share the same signature. Each domain contains 10 functions, covering arithmetic operations, string manipulation, list aggregation, and Boolean logic across the full set.

\begin{table*}[t]
\centering
\small
\setlength{\tabcolsep}{5pt}
\renewcommand{\arraystretch}{1.25}
\begin{tabular}{@{}>{\raggedright\arraybackslash}p{0.11\textwidth} >{\raggedright\arraybackslash}p{0.21\textwidth} >{\raggedright\arraybackslash}p{0.62\textwidth}@{}}
\toprule
\textbf{Domain} & \textbf{Signature} & \textbf{Target functions} \\
\midrule
Number & \texttt{(x: int) -> int} & \texttt{difference\_from\_reversed\_absolute}, \texttt{smallest\_divisor\_ge\_two\_abs}, \texttt{distance\_from\_square\_of\_three}, \texttt{add\_seven\_if\_negative}, \texttt{max\_with\_negative\_self}, \texttt{times\_four\_mod\_nine}, \texttt{integer\_average\_with\_ten}, \texttt{modulo\_of\_cube\_by\_eleven}, \texttt{fibonacci\_index\_small}, \texttt{absolute\_modulus\_gap\_eleven} \\
\midrule
Number Pairs & \texttt{(a: int, b: int) -> int} & \texttt{last\_digit\_sum}, \texttt{zero\_if\_opposite\_else\_product}, \texttt{absolute\_sum\_minus\_absolute\_difference}, \texttt{sum\_if\_negative}, \texttt{difference\_squared}, \texttt{sum\_after\_incrementing\_first}, \texttt{manhattan\_to\_pair}, \texttt{sum\_plus\_larger\_abs}, \texttt{triple\_sum\_minus\_product}, \texttt{product\_mod\_seven} \\
\midrule
String & \texttt{(text: str) -> str} & \texttt{rot13\_lowercase\_only}, \texttt{keep\_whitespace\_only}, \texttt{mirror\_with\_pipe}, \texttt{strip\_and\_lowercase}, \texttt{repeat\_three\_times}, \texttt{remove\_whitespace}, \texttt{hex\_code\_points}, \texttt{take\_last\_three}, \texttt{letters\_only}, \texttt{prepend\_hash} \\
\midrule
List & \texttt{(items: List[int]) -> int} & \texttt{sum\_excluding\_max}, \texttt{sum\_mod\_three}, \texttt{count\_adjacent\_opposite\_pairs}, \texttt{sum\_smaller\_of\_neighbors}, \texttt{count\_distinct\_values}, \texttt{sum\_of\_positive\_squares}, \texttt{sum\_negative\_values}, \texttt{sum\_neighbors\_products}, \texttt{sum\_palindrome\_values}, \texttt{count\_odd\_values} \\
\midrule
Logic & \texttt{(a: bool, b: bool) -> bool} & \texttt{both\_are\_true\_identity}, \texttt{b\_without\_a}, \texttt{a\_requires\_b}, \texttt{second\_is\_majority}, \texttt{bool\_from\_any\_tuple}, \texttt{a\_or\_b\_via\_if}, \texttt{difference\_negative}, \texttt{nor\_result}, \texttt{all\_or\_none}, \texttt{reverse\_implication} \\
\bottomrule
\end{tabular}
\caption{The 50 target functions of Alien Abduction: for each domain, 10 functions randomly sampled from the validated, GPT5.4-generated pool (Section~\ref{sec:target_functions}).}
\label{tab:alien_abduction_functions}
\end{table*}

\section{Prompt Templates}
\label{sec:function-detective-prompts}
Following standard prompting approaches~\citep{DBLP:conf/nips/BrownMRSKDNSSAA20, DBLP:conf/nips/Wei0SBIXCLZ22, DBLP:journals/csur/LiuYFJHN23}, we use zero-shot prompts for all the multi-turn mode experiments. The prompts vary by who controls the evidence and its form, but follow a typical structure: task description, constraints, response format restrictions. Figure~\ref{fig:mode_prompts} shows the condensed initial prompt for each of the six game modes, instantiated for a Number domain target with signature \texttt{(x: int) -> int}. The \textit{Setup} line contains the mode identifier used in the accompanying code and instance files. In our preliminary experiments, we used 15 turns for the multi-turn interaction modes and a maximum of 300 tokens. These limits worked well without any truncation or aborted runs and we therefore use the same settings across all experiments.

\begin{table*}[]
\centering
\begin{tabular}{cccc}
\hline
Model           & \% of Instances Failed & \makecell{\% Failures that are \\ Not Committed} & \makecell{\% Not Committed and \\ 100\% HRA at Last Turn}  \\ \hline
\verb|Qwen3.6-35B|     & 97.00  & 94.80 & 10.20 \\
\verb|Mistral-Large-3| & 80.30  & 46.50 & 0.00  \\
\verb|GPT5.4-Mini|     & 66.00  & 14.10 & \textbf{53.60}\\
\verb|GPT5.4|          & 35.70  & 0.90  & 0.0 \\ \hline
\end{tabular}%
\caption{Failure and non-commitment rates across models. The final column reports uncommitted failures with an HRA of 1 at the last turn.}
\label{tab:failstats}
\end{table*}

\begin{table*}
\centering
\begin{tabular}{cccc}
\hline
Model  & \makecell{Total Turns \\ (across 300 instances)} & Median Turns & \makecell{\% of Turns with \\ parser errors} \\ \hline
\verb|Qwen3.6-35B|     & 2965  & 15  & 40.60 \\
\verb|Mistral-Large-3| & 2261  & 7.5 & 45.20\\
\verb|GPT5.4-Mini|     & 1379  & 3.0 & 20.40\\
\verb|GPT5.4|          & 1616  & 4.0 & 4.60\\ \hline
\end{tabular}%
\caption{Turn usage and parser-error rates across models over 300 instances.}
\label{tab:parsererrors}
\end{table*}

\section{Quantitative Analysis}
\label{sec:appendix-quantanalysis}
We used an AI-assisted code generator for the Matplotlib visualisation scripts. The authors reviewed, modified, and verified all generated code and figures.

\paragraph{Task Success Rates} We report task success rates across domains for all evaluated models. As shown in Figures~\ref{fig:successrate-1}--~\ref{fig:successrate-3}, Qwen3.6-35B obtains zero success in several domains and modes. As discussed in Section~\ref{subsec:evidence-budget}, the model often exhausts the turn budget without committing to a solution. In active-output, it achieves non-zero success rates for Number, String, and Logic, at 0.1, 0.2, and 0.4, respectively. Its success rate is zero in both single-turn modes because the responses result in parser errors.

GPT5.4 obtains zero success for String in active-verdict, but reaches a success rate of 1.0 for Logic in both single-turn-output and passive-output. Its success rates in active-verdict are generally lower across domains, except for List, where it scores 0.6. Overall, the domain-level results show variation across models and interaction modes, with Qwen3.6-35B particularly affected by non-commitment and parser errors, and GPT5.4 performing weakest in active-verdict.

\paragraph{Parser Errors} Table~\ref{tab:parsererrors} reports the frequency of model responses that do not follow the required format. Parser errors are most frequent for Mistral-Large-3 and Qwen3.6-35B, affecting 45.2\% and 40.6\% of their turns, respectively. GPT5.4-mini has a lower rate of 20.4\%, while GPT5.4 has the lowest rate at 4.6\%. These differences indicate that some models have greater difficulty following the structured interaction protocol. Moreover, frequent parser errors reduce the number of valid turns available within the turn budget and may contribute to non-commitment. Their effect is particularly visible in the single-turn modes for Qwen3.6-35B, where an unparsable response results in failure because the model has only one turn to submit a solution.

\paragraph{Hypothesis Retrodiction Accuracy} Figures~\ref{fig:hrarate-1}--~\ref{fig:hrarate-3} report average HRA across domains. Missing bars indicate that no valid HRA value was available for that setting. Active-output produces consistently high HRA across the available models, particularly for Number, String, and Logic. Passive-output is more variable: HRA remains high for Number Pairs and Logic but is lower for Numbers, String and List. Verdict-based modes also show greater variation across domains, with lower HRA in passive-verdict for Number Pairs, List, and Logic. Overall, these results show that the aggregate pattern is not uniform across domains.

\section{Qualitative Analysis}
\label{sec:appendix-qualitative-analysis}
\paragraph{Negative Evidence Impact on Task Success} Figures~\ref{fig:negative-list-gpt54}, \ref{fig:negative-logic-gpt54}, and~\ref{fig:negative-twonumbers-gpt54} compare how models respond to negative evidence in active-verdict and passive-verdict. As discussed in Section~\ref{sec:qualitativeanalysis}, in active-verdict, each negative verdict rejects only a model-proposed input--output pair. For List and Number Pairs, several failed instances accumulate negative verdicts while the reported hypotheses remain inconsistent with the observed evidence or unknown, and the interaction often continues close to the turn limit. In passive-verdict, the Game Master provides both positive and negative examples, and models generally reach evidence-consistent hypotheses in fewer turns, although some failures remain. The difference is less pronounced for Logic, where interactions in both modes usually end within a few turns. Overall, the traces suggest that negative evidence is more useful when it is provided alongside broader example coverage than when it only rejects self-selected proposals.
\paragraph{Confidence Score} Figure~\ref{fig:lastturn-confidence} compares the models' self-reported confidence on successful and failed instances. Confidence is generally higher for successful instances, but the separation varies across models and modes. Qwen3.6-35B shows gaps in active-output and passive-output, while GPT5.4-mini shows clearer separation in the passive and single-turn modes. In contrast, Mistral-Large-3 remains highly confident even on failed instances, with little or no separation in the single-turn modes. GPT5.4 also shows similar confidence for successful and failed instances in passive-verdict. These results suggest that confidence does not always reliably distinguish correct from incorrect final hypotheses. Missing bars indicate that no instances were available for the corresponding outcome.
\paragraph{Query Coverage} Figures~\ref{fig:querycoverage-list}, ~\ref{fig:querycoverage-logic} and~\ref{fig:querycoverage-numbers} show the query coverage of GPT5.4 across domains. As discussed in Section~\ref{subsec:queryselection}, the self-selected queries in the active modes cover a narrow range than in the passive modes. Overall, this suggests that GPT5.4's self-selected queries explore only a limited portion of the input space.

\paragraph{Current Interaction State} Although the current interaction state is self-reported by models, it may not completely reflect their internal belief states. We therefore use this information only as complementary evidence for understanding the model behavior. We hypothesize that a model that reliably maintains a belief state would indicate a confirming state before committing to a solution. Figure~\ref{fig:lastturn-interactionstate} shows the interaction states reported by models at the last turn. Qwen3.6-35B's responses mostly indicate a \textit{probing} state rather than \textit{confirming}. This is consistent with the findings in Section~\ref{subsec:modelsuccessanalysis}, where the model often exhausts its turn budget. On the other hand, GPT5.4 indicates confirming in most modes, while Mistral-Large-3 does so in several modes. GPT5.4-mini sometimes uses \textit{Proposing} or \textit{Solving}, which do not match the response choices given in the prompt. Moreover, in both active-output and active-verdict, the model indicates probing at the last turn and then immediately commits to a solution, implying that the model may not be accurately reporting its internal belief state.

%\FloatBarrier
% Prompt templates for the six game modes. Originally auto-generated from
% Alien-Abduction prompt sources; manually curated for the paper.

\begin{figure*}[t]
\centering
  \vspace{-0.3cm}  
  \includegraphics[width=0.80\textwidth]{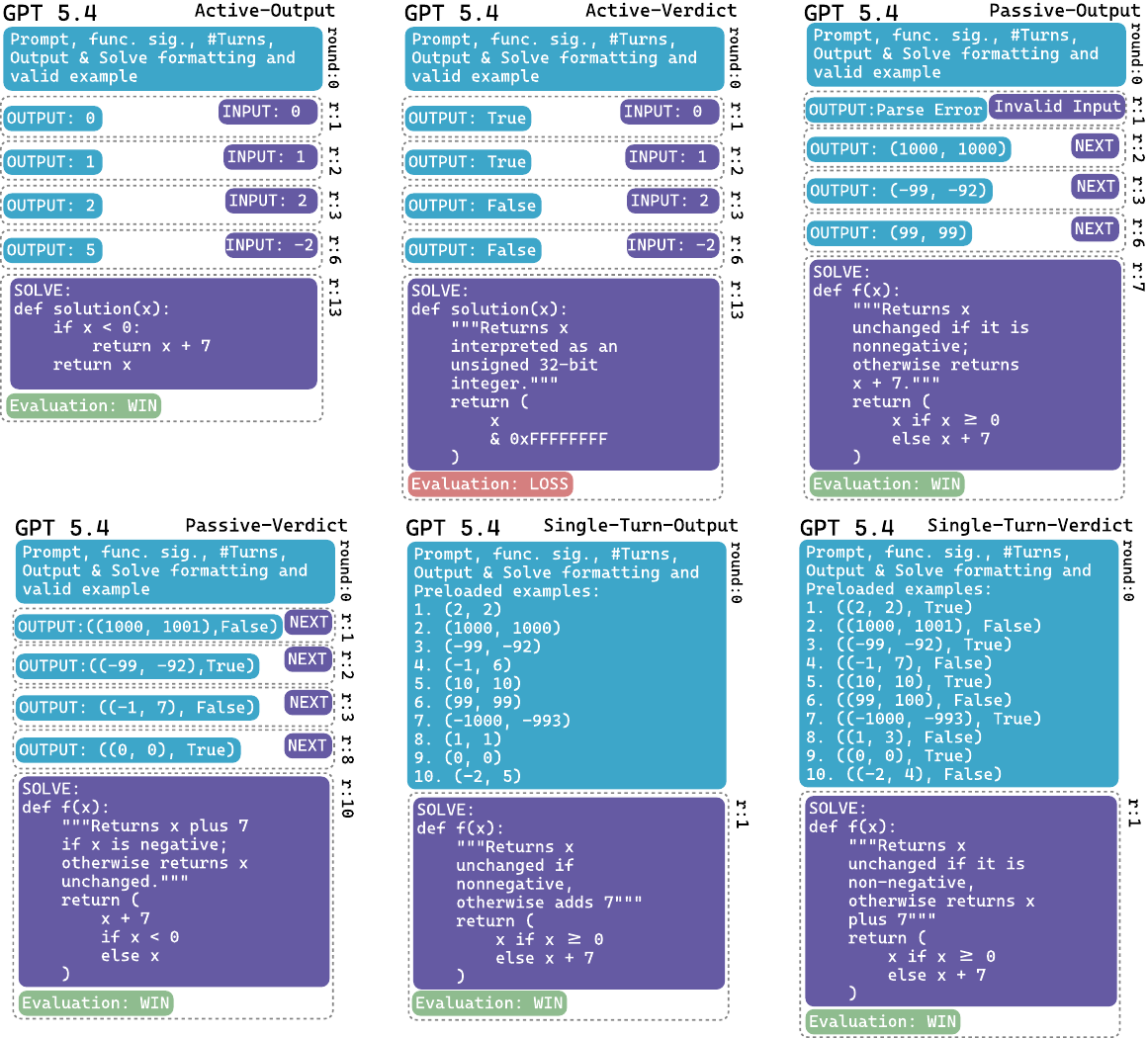}
  \caption{Qualitative analysis of GPT-5.4 on all variants. The chosen actions and corresponding outputs are given across turns. Initial prompt that covers the tasks, the given function signature is excluded for visualisation reasons.}
  \label{fig:qual_analysis1}
  \vspace{-0.3cm}  
\end{figure*}

\begin{figure*}[t]
    \centering
   \includegraphics[width=0.95\linewidth]{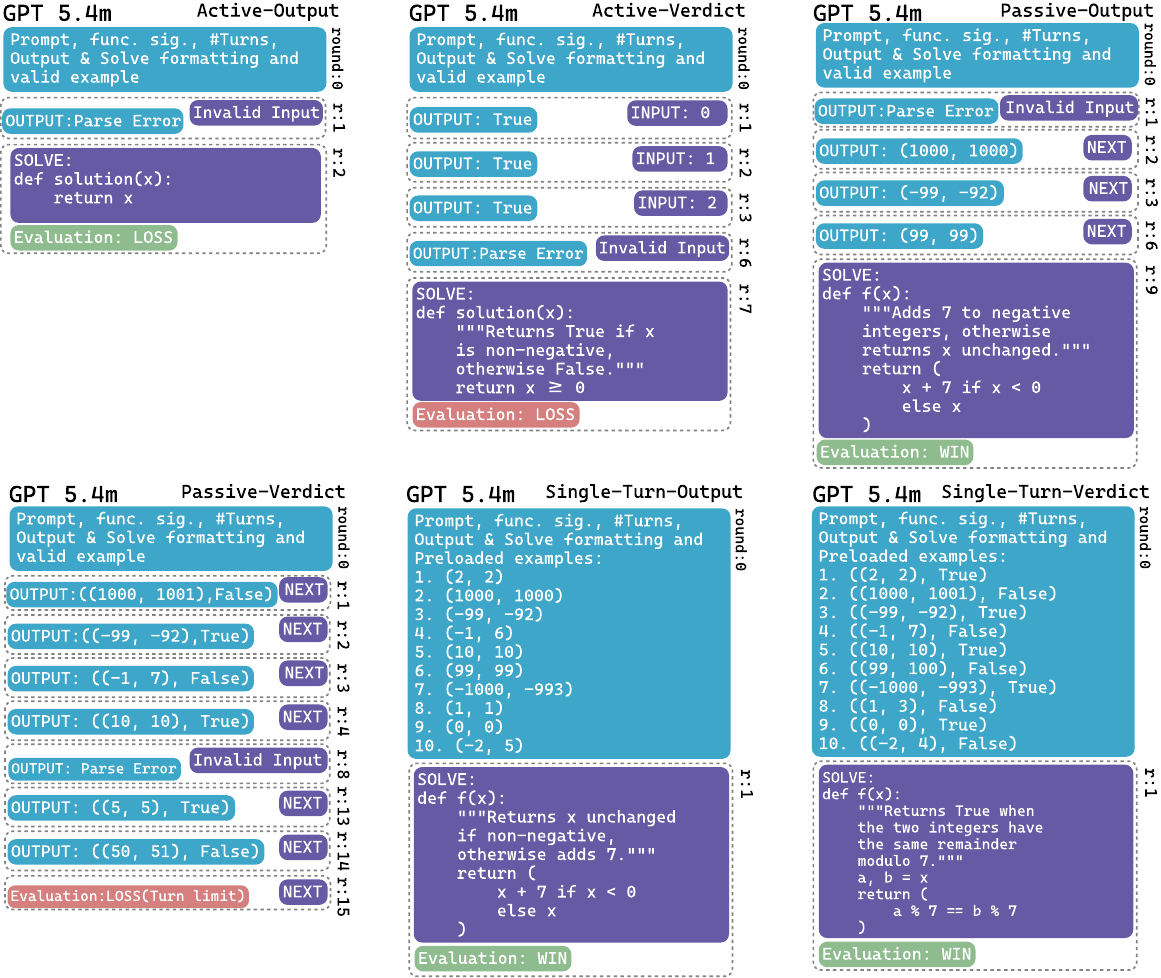}
    \caption{GPT-5.4-mini outputs for six variants}
    \label{fig:qual_analysis2}    
\end{figure*}

\begin{figure*}[t]
    \centering
   \includegraphics[width=0.95\linewidth]{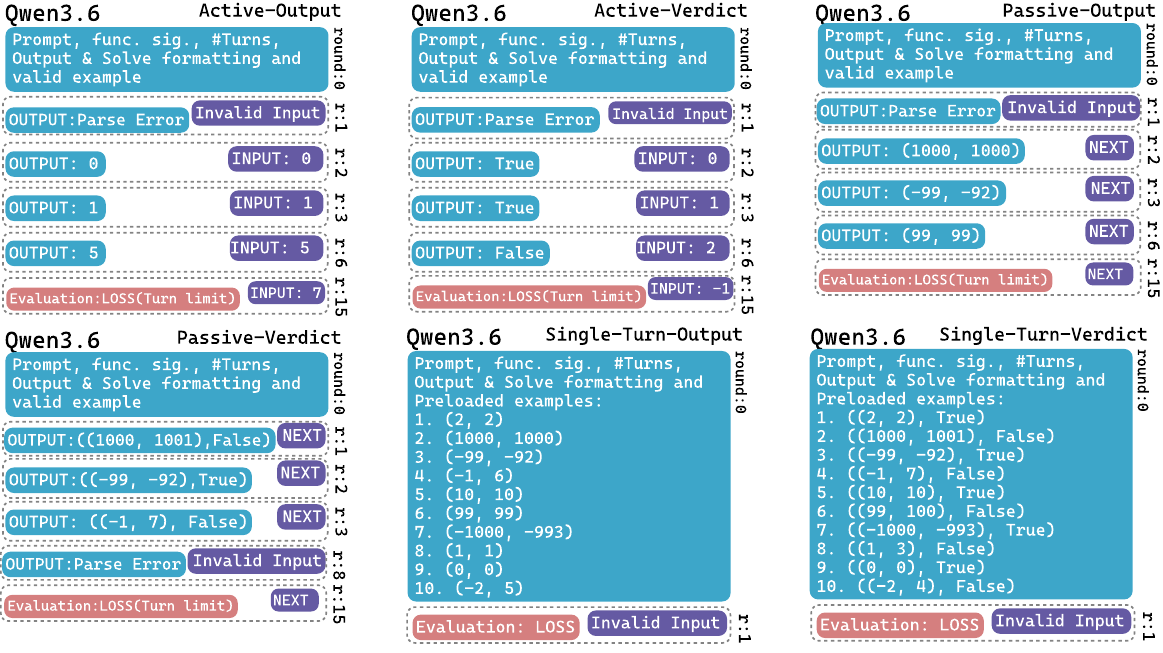}
    \caption{Qwen-3.6 outputs for six variants}
    \label{fig:qual_analysis3}    
\end{figure*}

\begin{figure*}[t]
    \centering
   \includegraphics[width=0.95\linewidth]{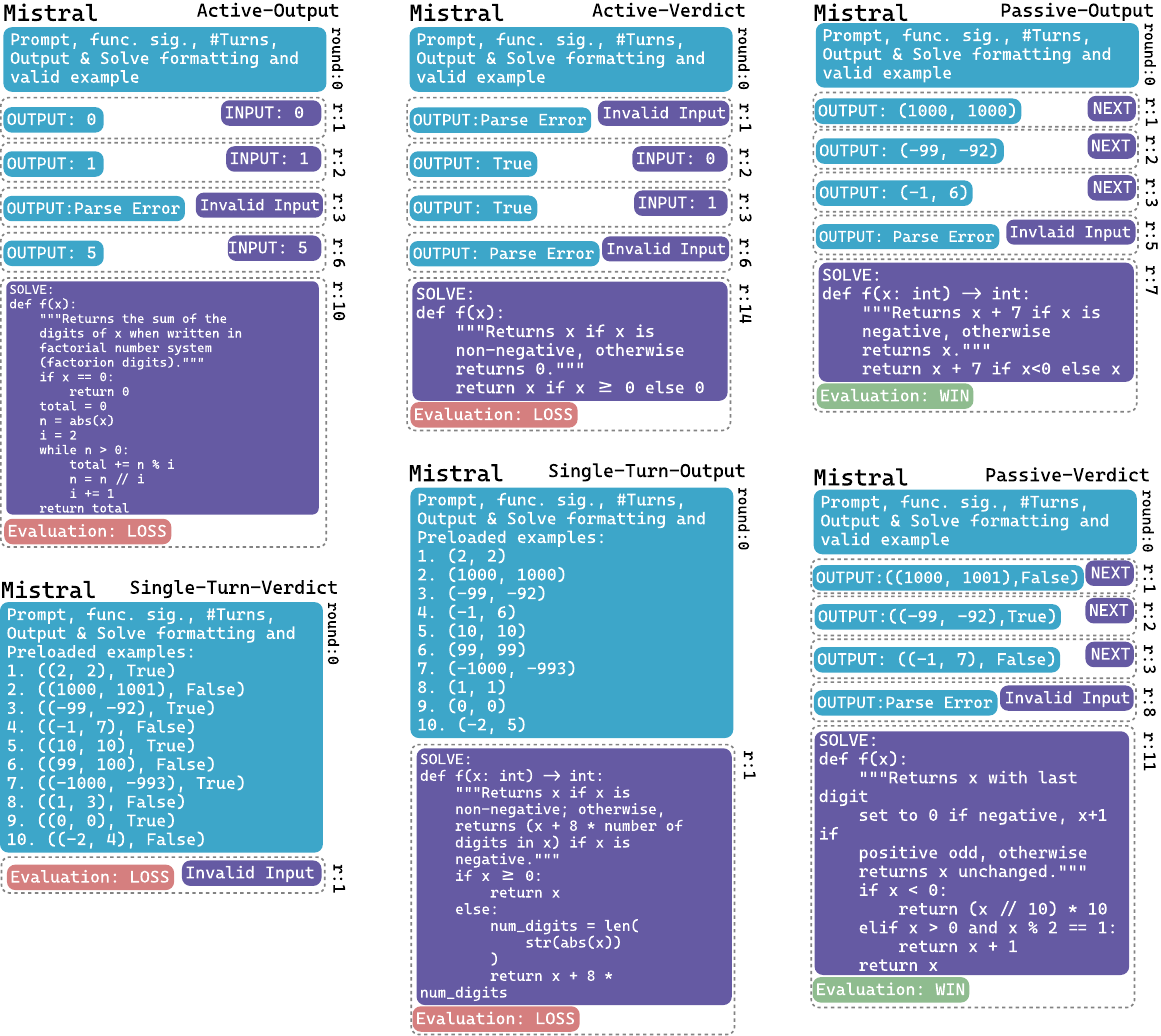}
    \caption{Mistral outputs for six variants}
    \label{fig:qual_analysis4}    
\end{figure*}

\begin{figure*}[t]
\begin{minipage}[t]{0.48\textwidth}
\begin{fdprompt}{Active-Output (AO)}
Infer hidden Python function from limited evidence.\\
Signature: (x: int) -> int\\
Turn budget: 15\\
Setup: Active Inputs\\
Action per turn: TEST: <inputs> or\\
\phantom{Action per turn: }SOLVE: ```python ... ```\\
GM reply format: OUTPUT: <value>\\
Visible example format: (-10, 10)\\
\end{fdprompt}
\end{minipage}\hfill
\begin{minipage}[t]{0.48\textwidth}
\begin{fdprompt}{Active-Verdict (AV)}
Infer hidden Python function from limited evidence.\\
Signature: (x: int) -> int\\
Turn budget: 15\\
Setup: Active Pair Checks\\
Action per turn: TEST: <inputs>, <candidate\_output>\\
\phantom{Action per turn: }or SOLVE: ```python ... ```\\
GM reply format: OUTPUT: True/False\\
Visible example format: ((-10, 10), True)\\
\end{fdprompt}
\end{minipage}

\begin{minipage}[t]{0.48\textwidth}
\begin{fdprompt}{Passive-Output (PO)}
Infer hidden Python function from limited evidence.\\
Signature: (x: int) -> int\\
Turn budget: 15\\
Setup: Passive Examples\\
Action per turn: NEXT: or SOLVE: ```python ... ```\\
GM reveals one valid pair at a time: (-10, 10)\\
GM reply format: OUTPUT: (<input>, <output>)\\
\end{fdprompt}
\end{minipage}\hfill
\begin{minipage}[t]{0.48\textwidth}
\begin{fdprompt}{Passive-Verdict (PV)}
Infer hidden Python function from limited evidence.\\
Signature: (x: int) -> int\\
Turn budget: 15\\
Setup: Passive Labeled Pairs\\
Action per turn: NEXT: or SOLVE: ```python ... ```\\
GM reveals one labeled pair at a time: ((-10, 10), True)\\
GM reply format:\\
\phantom{GM reply f}OUTPUT: ((<input>, <candidate\_output>), True/False)\\
\end{fdprompt}
\end{minipage}

\begin{minipage}[t]{0.48\textwidth}
\begin{fdprompt}{Single-Turn-Output (STO)}
Infer hidden Python function from limited evidence.\\
Signature: (x: int) -> int\\
Turn budget: 1\\
Setup: Passive Examples One-Shot\\
Only action: SOLVE: ```python ... ```\\
All valid examples are shown up front;\\
no NEXT requests allowed.\\
Preloaded batch: (-10, 10) | (0, 0) ...\\
\end{fdprompt}
\end{minipage}\hfill
\begin{minipage}[t]{0.48\textwidth}
\begin{fdprompt}{Single-Turn-Verdict (STV)}
Infer hidden Python function from limited evidence.\\
Signature: (x: int) -> int\\
Turn budget: 1\\
Setup: Passive Labeled Pairs One-Shot\\
Only action: SOLVE: ```python ... ```\\
All labeled candidate pairs are shown up front;\\
no NEXT requests allowed.\\
Preloaded batch: ((-10, 10), True) | ((0, -1), False) ...\\
\end{fdprompt}
\end{minipage}
\caption{Condensed initial prompts for the six game modes of Alien Abduction, shown for a Number domain target with signature \texttt{(x: int) -> int}.}
\label{fig:mode_prompts}
\end{figure*}

% Example episodes from Alien-Abduction transcripts. Originally auto-generated;
% manually curated for the paper.

\section{Example Episodes}
\label{sec:function-detective-episodes}

This appendix shows two lightly shortened Single-Turn-Output episodes: one lost (Table~\ref{tab:episode_loss}) and one won (Table~\ref{tab:episode_win}). Game Master messages are shown in gray and player responses in green; the final row of each transcript shows the hidden target function.

\begin{table*}[t]
\centering
\footnotesize
\setlength{\tabcolsep}{4pt}
\begin{tabular}{c@{$\;$}|p{.46\textwidth}p{.46\textwidth}}
\# & \textbf{Game Master} & \textbf{Player} \\
\hline
1
& \cellcolor[rgb]{0.9,0.9,0.9}{
\makecell[{{p{\linewidth}}}]{
\texttt{Infer hidden Python function from fixed evidence.} \\
\texttt{You know: (x: int) -> int; You have 1 turns total.} \\
\texttt{Mode: Passive Examples One-Shot} \\
\texttt{One action only: SOLVE: ```python ... ```} \\
\texttt{...} \\
\texttt{Preloaded examples:} \\
\texttt{1. (-10, 10) | 2. (0, 0) | 3. (-2, 2) | 4. (-99, 99) | 5. (99, 99)} \\
\texttt{6. (10, 10) | 7. (1000, 1000) | 8. (2, 2) | 9. (-1, 1) | 10. (1, 1)} \\
}
}
&  \\ \\
2
&
& \cellcolor[rgb]{0.561,0.737,0.561}{
\makecell[{{p{\linewidth}}}]{
\texttt{SOLVE: ```python} \\
\texttt{def solution(x):} \\
\texttt{    return x} \\
\texttt{```} \\
}
} \\ \\
3
& \cellcolor[rgb]{0.9,0.9,0.9}{
\makecell[{{p{\linewidth}}}]{
\texttt{Game Verdict: LOSS} \\
}
}
&  \\ \\
4
& \cellcolor[rgb]{0.9,0.9,0.9}{
\makecell[{{p{\linewidth}}}]{
\texttt{Incorrect.} \\
\texttt{Failed Cases:} \\
\texttt{Input: [-10] | Expected: 10 | Got: -10} \\
\texttt{Input: [-2] | Expected: 2 | Got: -2} \\
\texttt{Input: [-99] | Expected: 99 | Got: -99} \\
}
}
&  \\ \\
5
& \cellcolor[rgb]{0.9,0.9,0.9}{
\makecell[{{p{\linewidth}}}]{
\texttt{def absolute\_value(x: int) -> int:} \\
\texttt{    return abs(x)} \\
}
}
&  \\ \\
\end{tabular}
\caption{A lost Single-Turn-Output episode. The player submits the identity function although the preloaded examples contain negative inputs with positive outputs; the hidden target is \texttt{absolute\_value}.}
\label{tab:episode_loss}
\end{table*}

\begin{table*}[t]
\centering
\footnotesize
\setlength{\tabcolsep}{4pt}
\begin{tabular}{c@{$\;$}|p{.46\textwidth}p{.46\textwidth}}
\# & \textbf{Game Master} & \textbf{Player} \\
\hline
1
& \cellcolor[rgb]{0.9,0.9,0.9}{
\makecell[{{p{\linewidth}}}]{
\texttt{Infer hidden Python function from fixed evidence.} \\
\texttt{You know: (text: str) -> str; You have 1 turns total.} \\
\texttt{Mode: Passive Examples One-Shot} \\
\texttt{One action only: SOLVE: ```python ... ```} \\
\texttt{...} \\
\texttt{Preloaded examples:} \\
\texttt{1. ('hello', 'h') | 2. ('hello world', 'h') | 3. ('123', '1') | 4. ('ab', 'a') | 5. ('a', 'a')} \\
\texttt{6. ('a-b', 'a') | 7. (' ', ' ') | 8. ('aba', 'a') | 9. ('   ', ' ') | 10. ('', '')} \\
}
}
&  \\ \\
2
&
& \cellcolor[rgb]{0.561,0.737,0.561}{
\makecell[{{p{\linewidth}}}]{
\texttt{SOLVE: ```python} \\
\texttt{def solution(text: str) -> str:} \\
\texttt{    return text[0] if text else ''} \\
\texttt{```} \\
}
} \\ \\
3
& \cellcolor[rgb]{0.9,0.9,0.9}{
\makecell[{{p{\linewidth}}}]{
\texttt{Game Verdict: WIN} \\
}
}
&  \\ \\
4
& \cellcolor[rgb]{0.9,0.9,0.9}{
\makecell[{{p{\linewidth}}}]{
\texttt{That is correct!} \\
}
}
&  \\ \\
5
& \cellcolor[rgb]{0.9,0.9,0.9}{
\makecell[{{p{\linewidth}}}]{
\texttt{def first\_char(text: str) -> str:} \\
\texttt{    return text[:1]} \\
}
}
&  \\ \\
\end{tabular}
\caption{A won Single-Turn-Output episode. The player infers the correct rule from the preloaded examples, including the empty-string edge case; the hidden target is \texttt{first\_char}.}
\label{tab:episode_win}
\end{table*}

\begin{figure*}[htbp]
    \centering
    \begin{subfigure}[t]{0.49\textwidth}
        \centering
        \includegraphics[width=\linewidth]{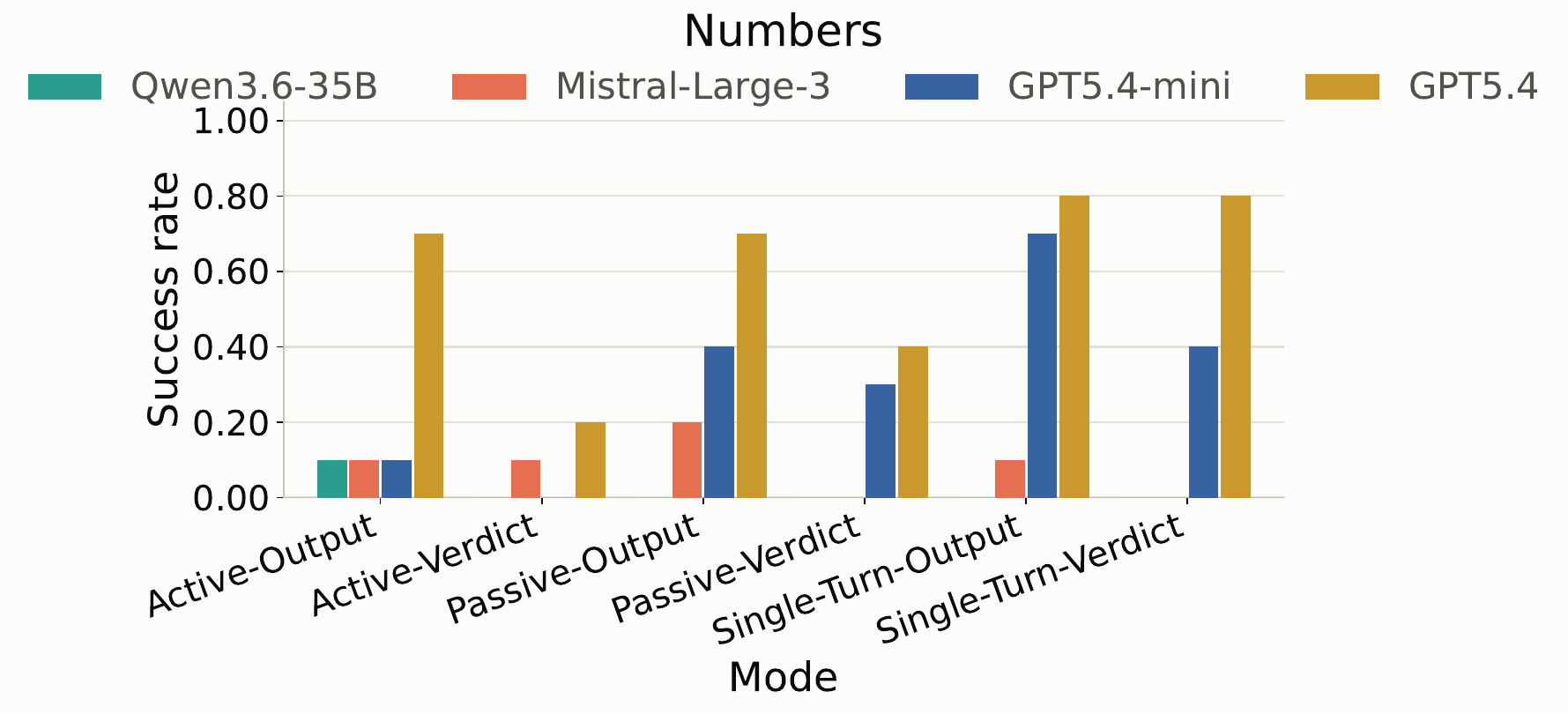}
        \caption{Task success rates for the Number domain.}
        \label{fig:successrate-numbers}
    \end{subfigure}
    \hfill
    \begin{subfigure}[t]{0.49\textwidth}
        \centering
        \includegraphics[width=\linewidth]{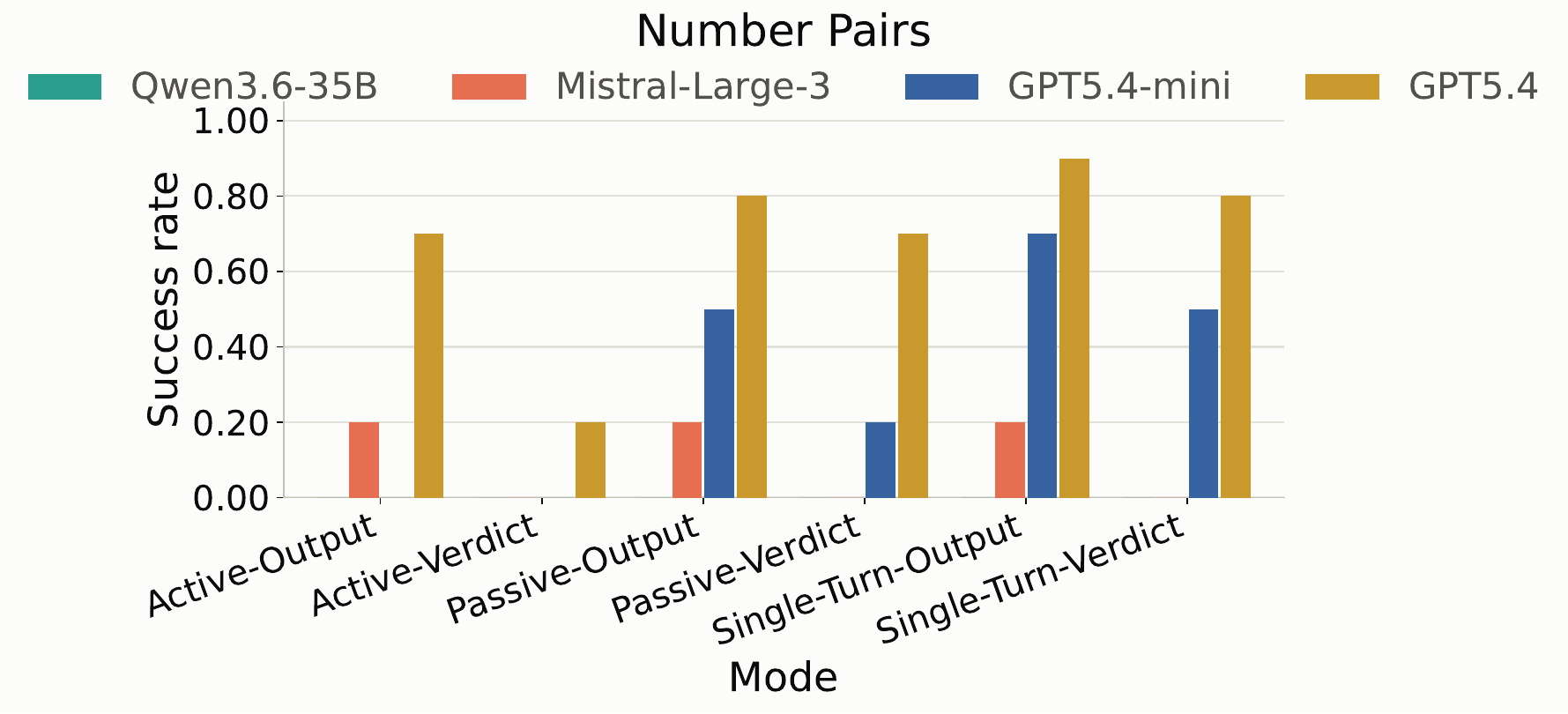}
        \caption{Task success rates for the Number Pairs domain.}
        \label{fig:successrate-numberpairs}
    \end{subfigure}
    \caption{Task success rates across models and interaction modes for the Number and Number Pairs domains. Missing bars indicate zero success.}
    \label{fig:successrate-1}
\end{figure*}

\begin{figure*}[htbp]
    \centering
    \hfill
    \begin{subfigure}[t]{0.49\textwidth}
        \centering
        \includegraphics[width=\linewidth]{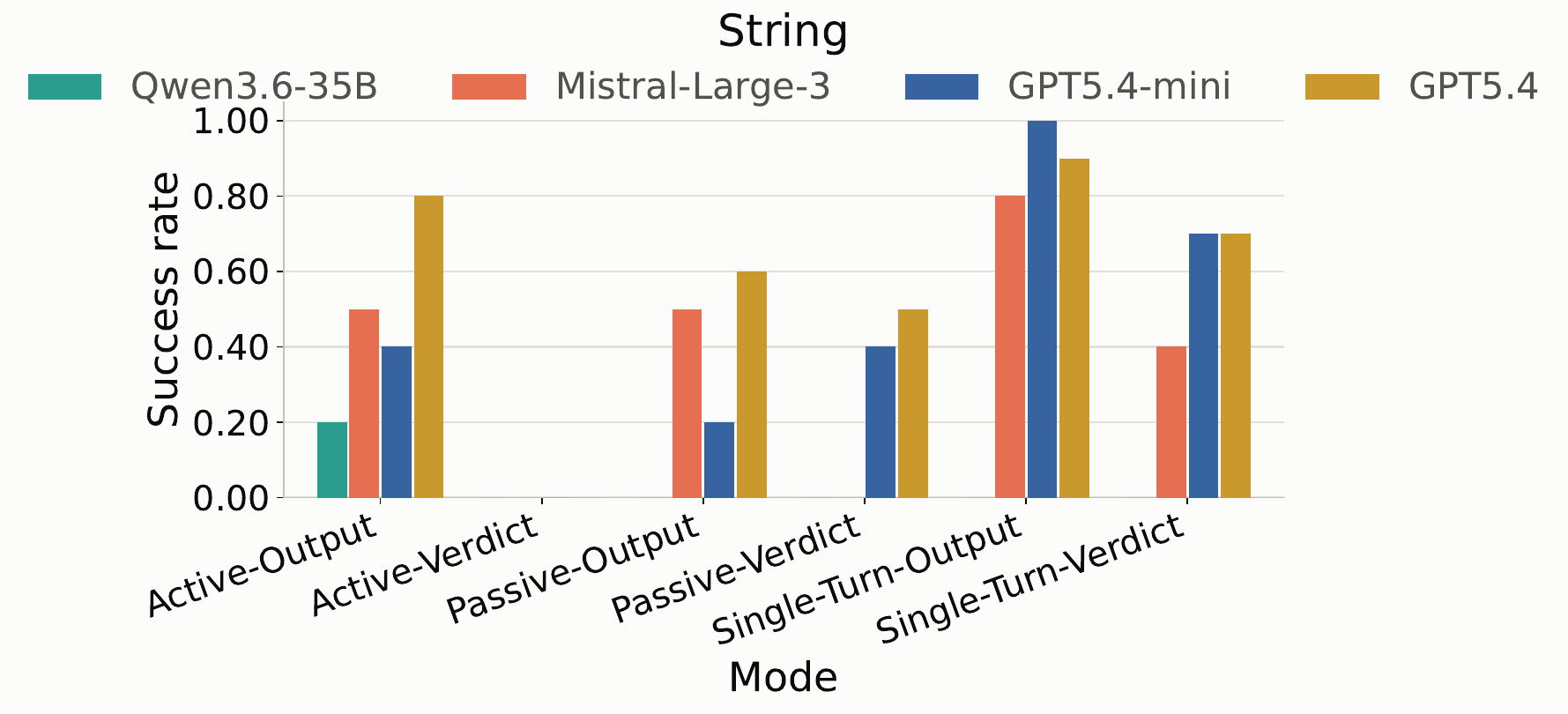}
        \caption{Task success rates for the String domain.}
        \label{fig:successrate-string}
    \end{subfigure}
    \hfill
    \begin{subfigure}[t]{0.49\textwidth}
        \centering
        \includegraphics[width=\linewidth]{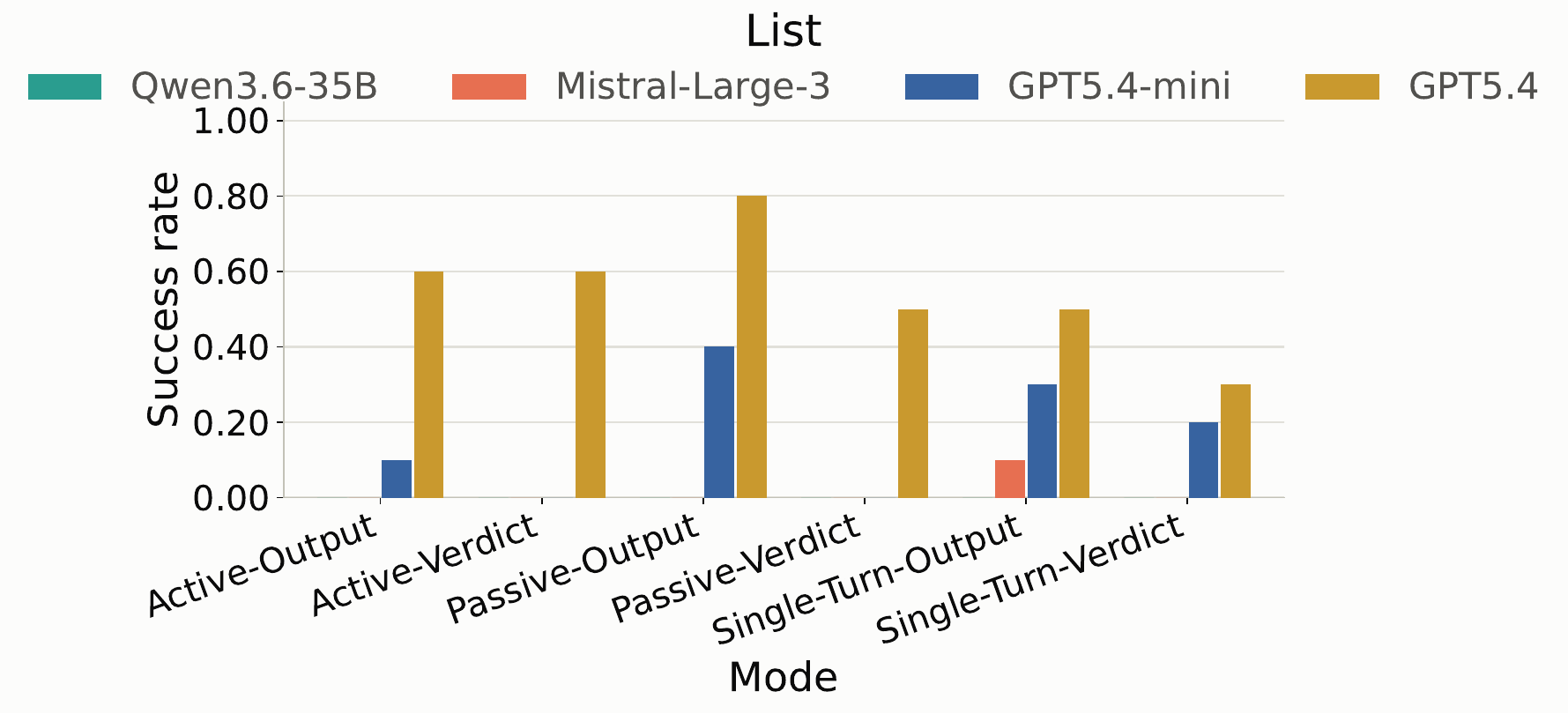}
        \caption{Task success rates for the List domain.}
        \label{fig:successrate-list}
    \end{subfigure}
    \caption{Task success rates across models and interaction modes for the String and List domains. Missing bars indicate zero success.}
    \label{fig:successrate-2}
\end{figure*}

\begin{figure*}[htbp]
    \centering
    \begin{subfigure}[t]{0.60\textwidth}
        \centering
        \includegraphics[width=\linewidth]{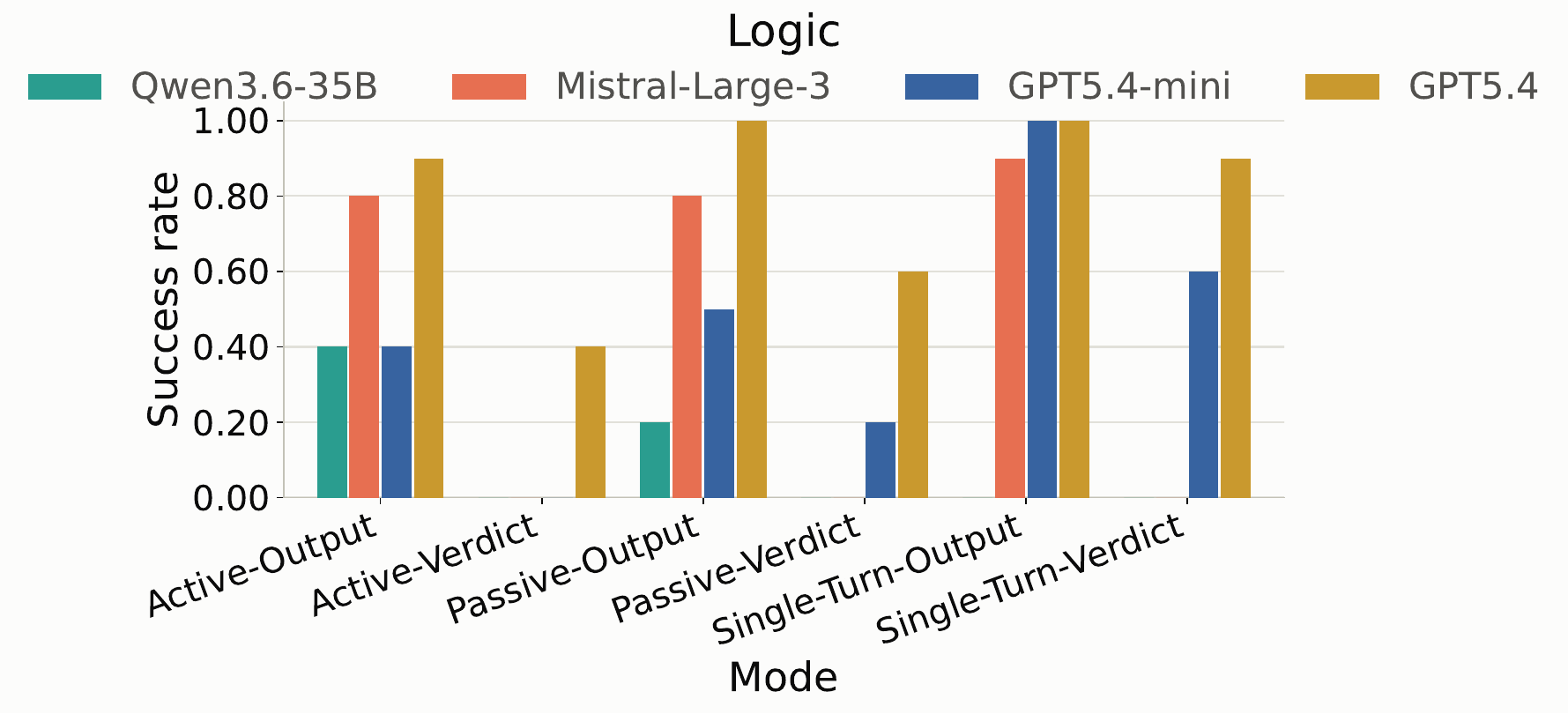}
        \label{fig:successrate-logic}
    \end{subfigure}
    \caption{Task success rates across models and interaction modes for the Logic domain. Missing bars indicate zero success.}
    \label{fig:successrate-3}
\end{figure*}

\begin{figure*}[htbp]
    \centering
    \begin{subfigure}[t]{0.49\textwidth}
        \centering
        \includegraphics[width=\linewidth]{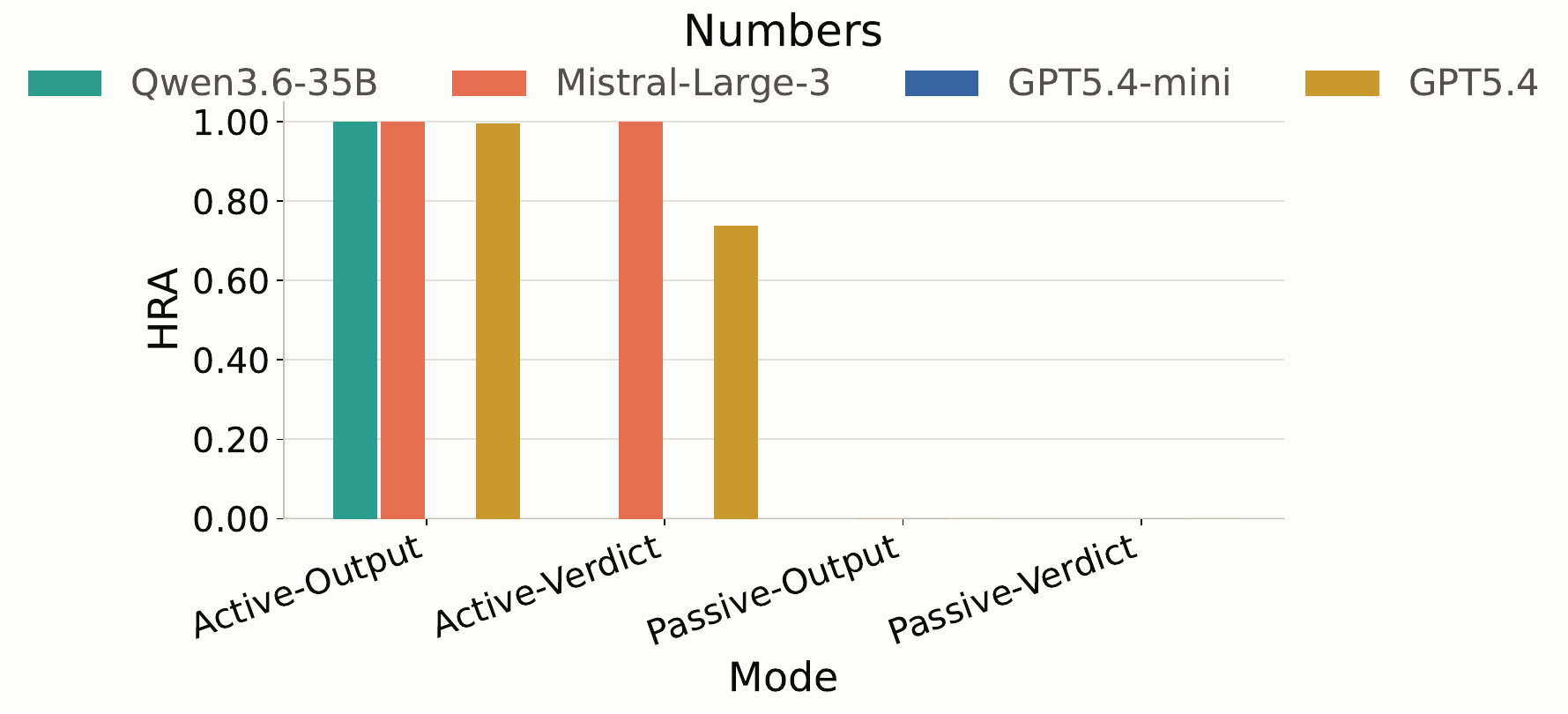}
        \caption{Hypothesis Retrodiction Accuracy for the Number domain.}
        \label{fig:hra-numbers}
    \end{subfigure}
    \hfill
    \begin{subfigure}[t]{0.49\textwidth}
        \centering
        \includegraphics[width=\linewidth]{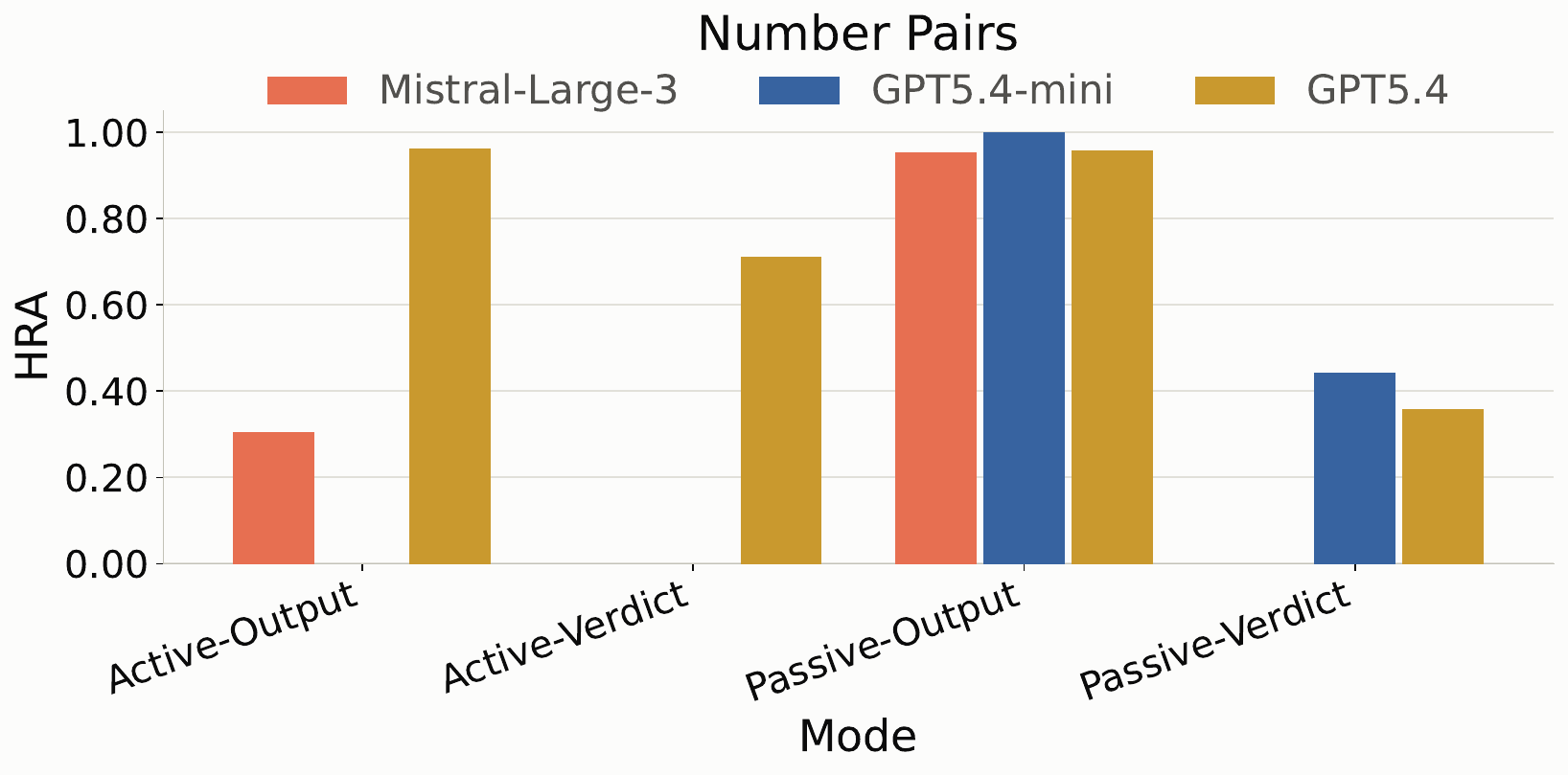}
        \caption{Hypothesis Retrodiction Accuracy for the Number Pairs domain.}
        \label{fig:hra-numberpairs}
    \end{subfigure}
    \caption{Hypothesis Retrodiction Accuracy across models and interaction modes for the Number and Number Pairs domains. Missing bars indicate zero success.}
    \label{fig:hrarate-1}
\end{figure*}

\begin{figure*}[htbp]
    \centering
    \hfill
    \begin{subfigure}[t]{0.49\textwidth}
        \centering
        \includegraphics[width=\linewidth]{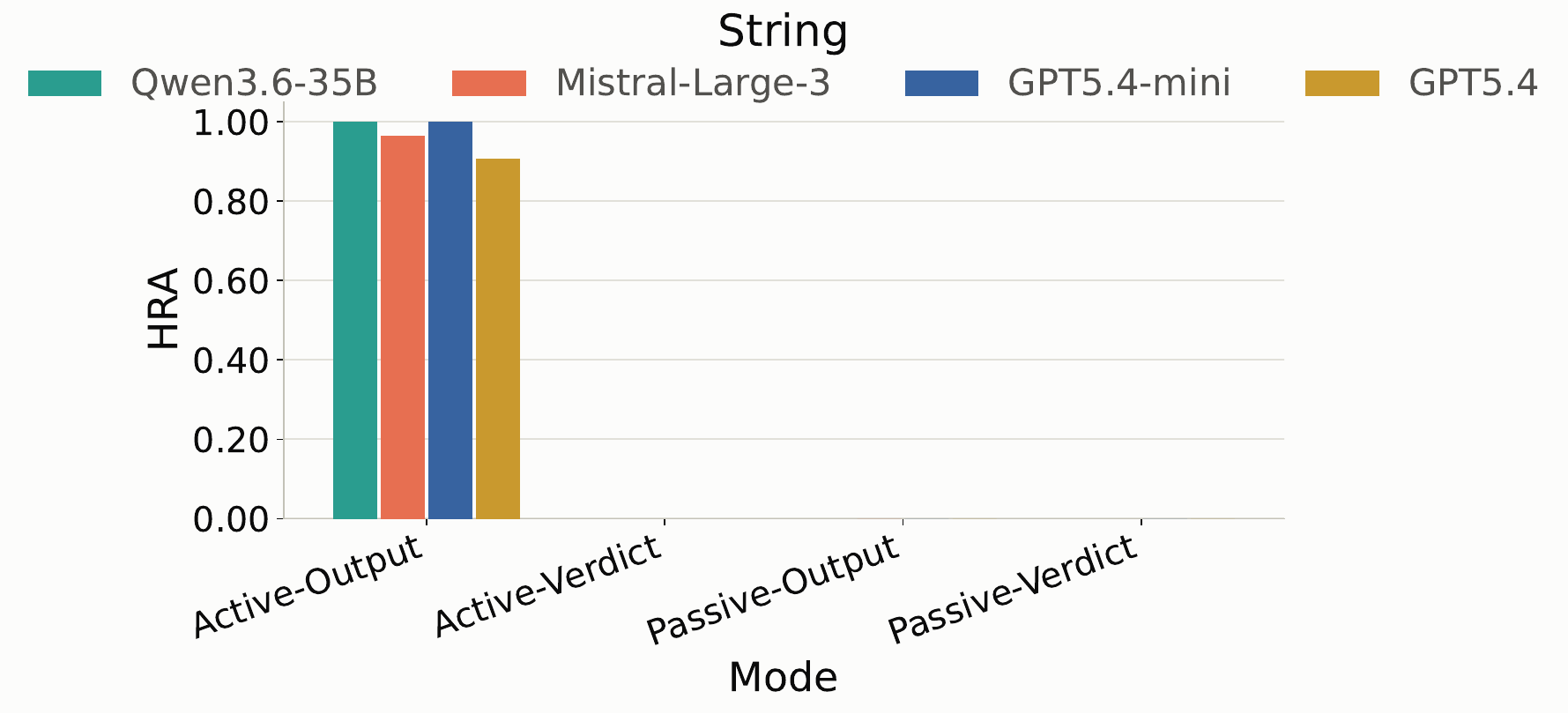}
        \caption{Hypothesis Retrodiction Accuracy for the String domain.}
        \label{fig:hra-string}
    \end{subfigure}
    \hfill
    \begin{subfigure}[t]{0.49\textwidth}
        \centering
        \includegraphics[width=\linewidth]{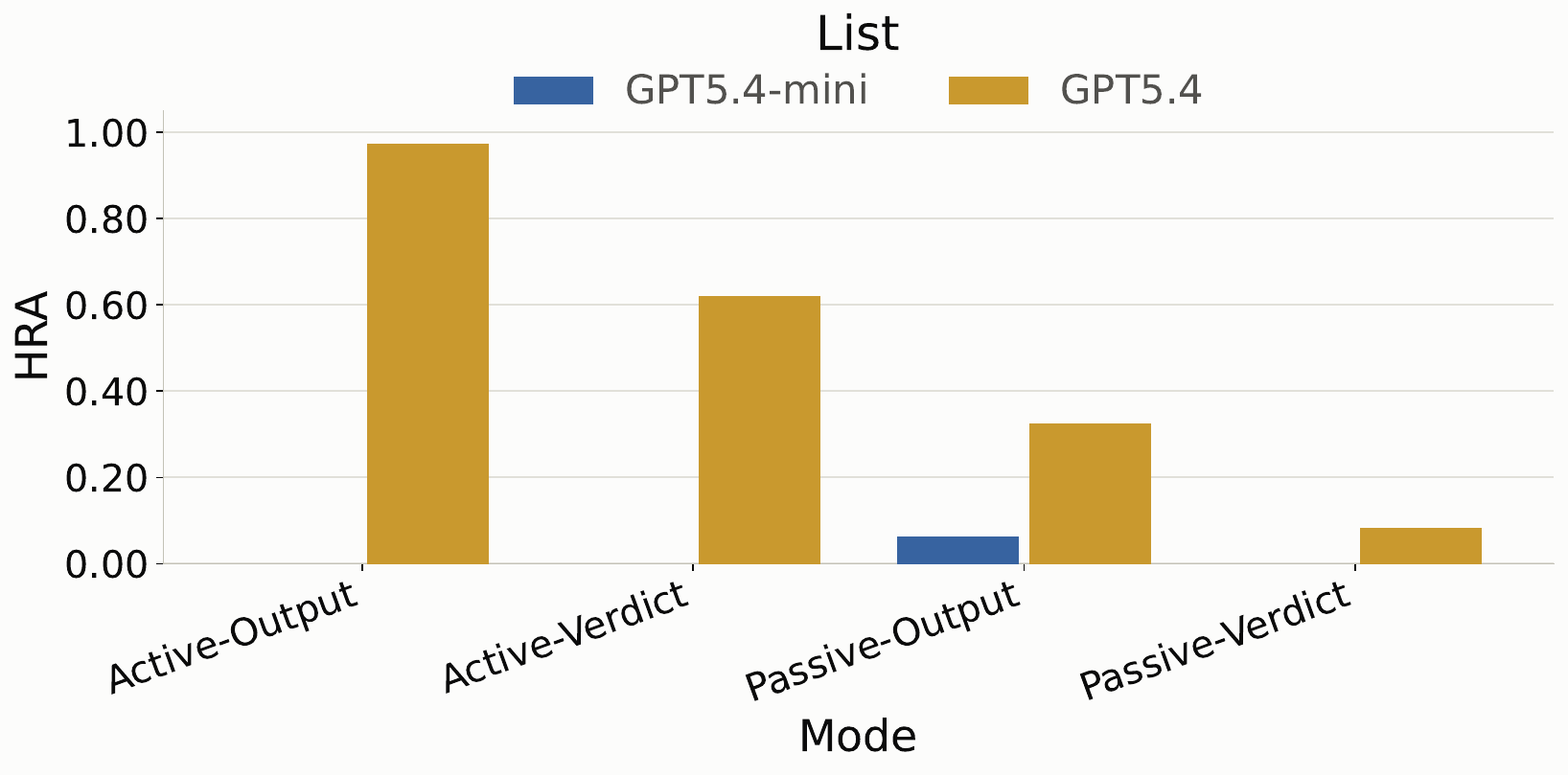}
        \caption{Hypothesis Retrodiction Accuracy for the List domain.}
        \label{fig:hra-list}
    \end{subfigure}
    \caption{Hypothesis Retrodiction Accuracy across models and interaction modes for the String and List domains. Missing bars indicate zero success.}
    \label{fig:hrarate-2}
\end{figure*}

\begin{figure*}[htbp]
    \centering
    \begin{subfigure}[t]{0.60\textwidth}
        \centering
        \includegraphics[width=\linewidth]{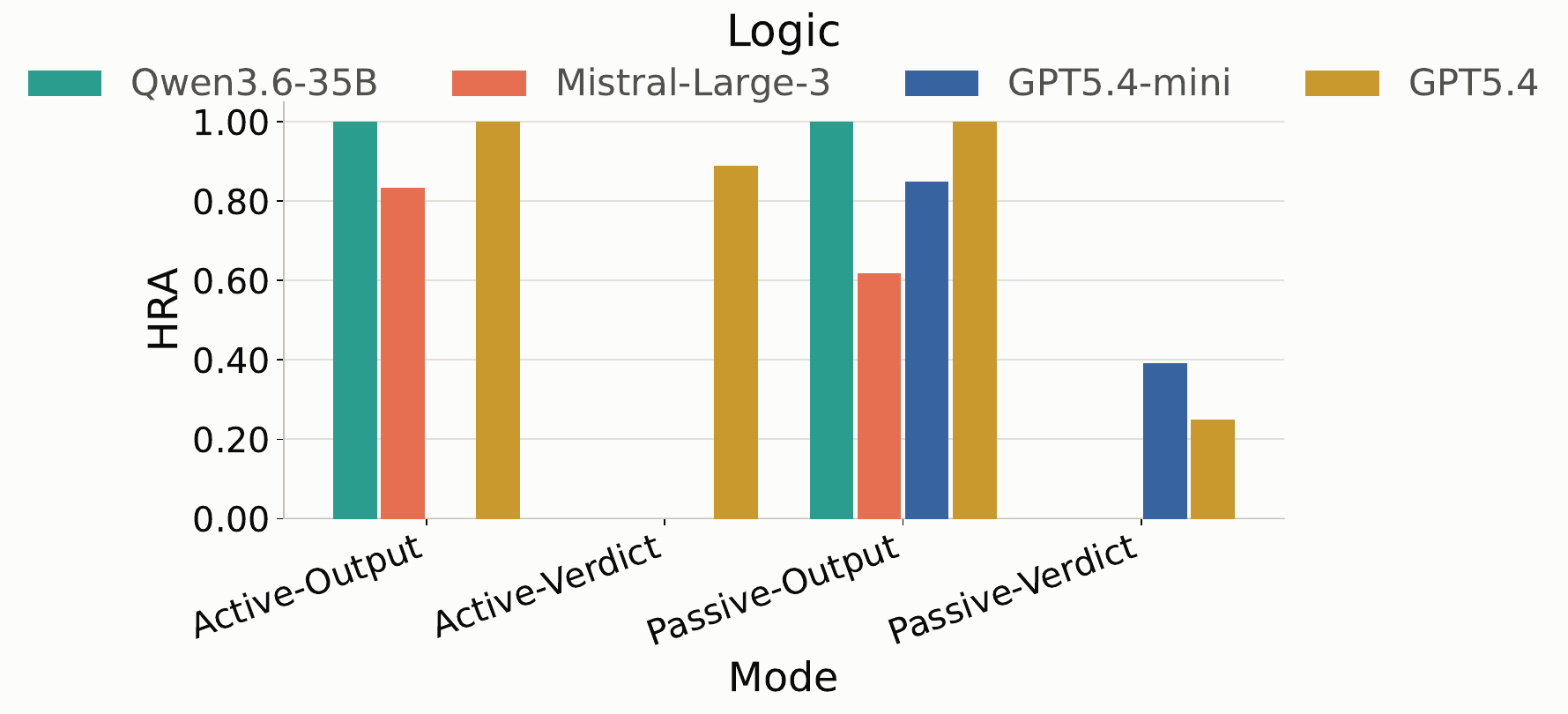}
        \label{fig:hra-logic}
    \end{subfigure}
    \caption{Hypothesis Retrodiction Accuracy across models and interaction modes for the Logic domain. Missing bars indicate zero success.}
    \label{fig:hrarate-3}
\end{figure*}

\begin{figure*}[t]
\centering
  \includegraphics[width=\textwidth]{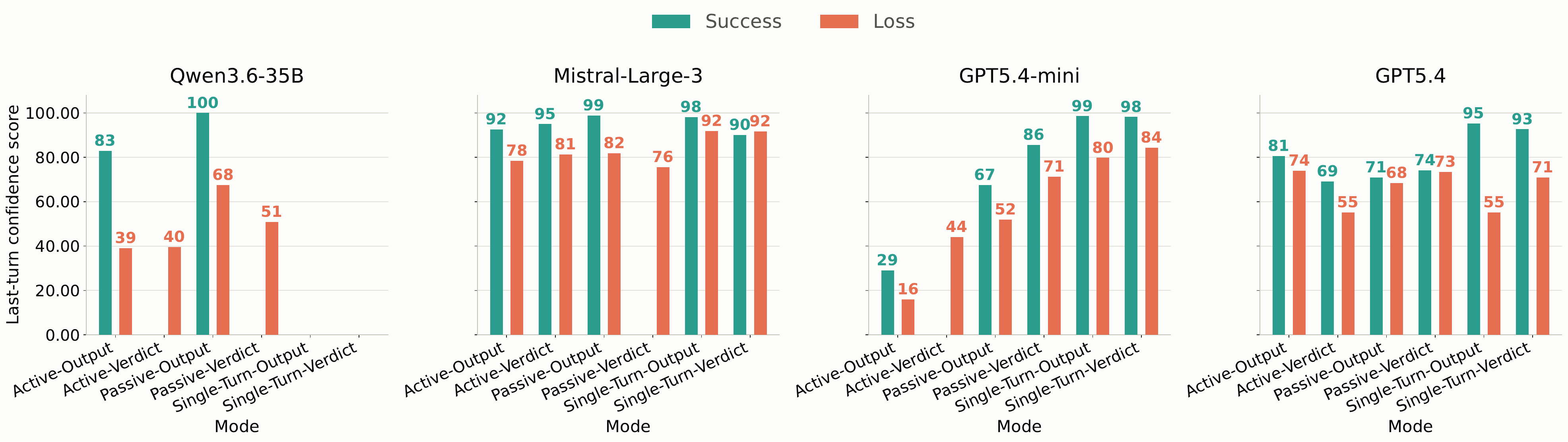}
  \caption{Confidence scores at the final turn across models and interaction modes.
  }
  \label{fig:lastturn-confidence}
  \vspace{-0.3cm}  
\end{figure*}

\begin{figure*}[t]
\centering
  \includegraphics[width=\textwidth]{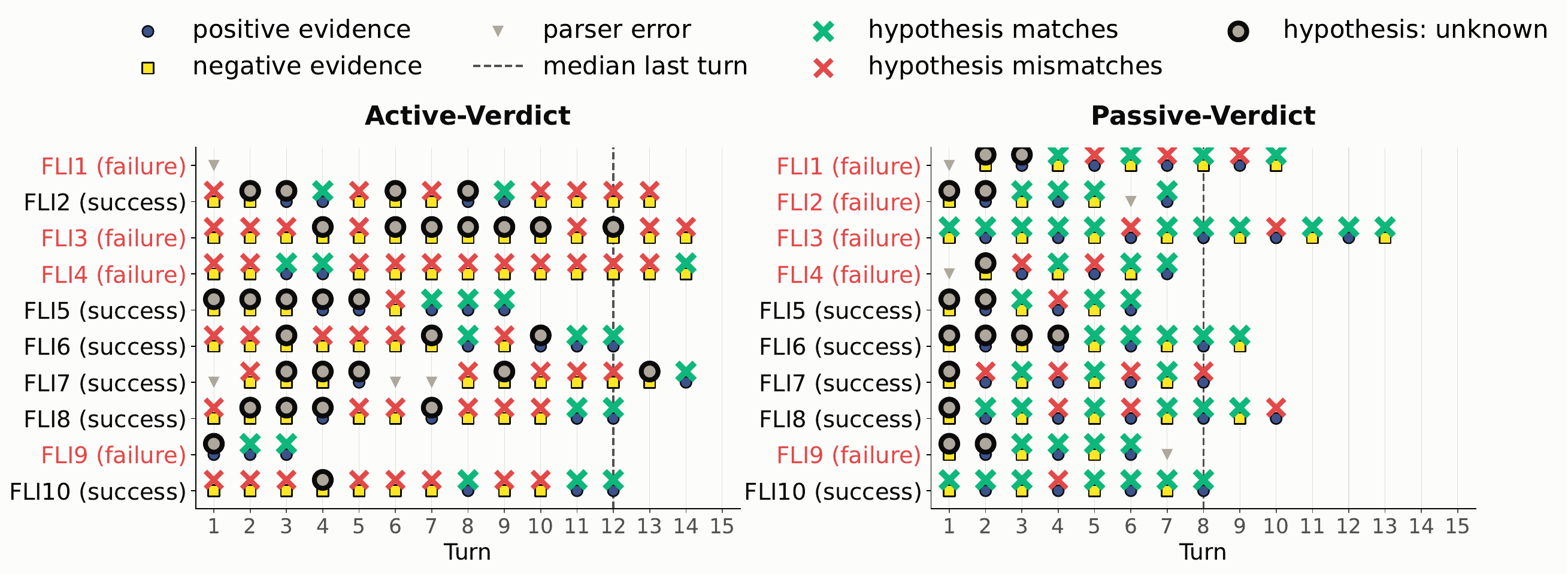}
  \caption{Negative evidence impact on task success across verdict modes 
  for GPT5.4 in List domain.
  }
  \label{fig:negative-list-gpt54}
  \vspace{-0.3cm}  
\end{figure*}

\begin{figure*}[t]
\centering
  \includegraphics[width=\textwidth]{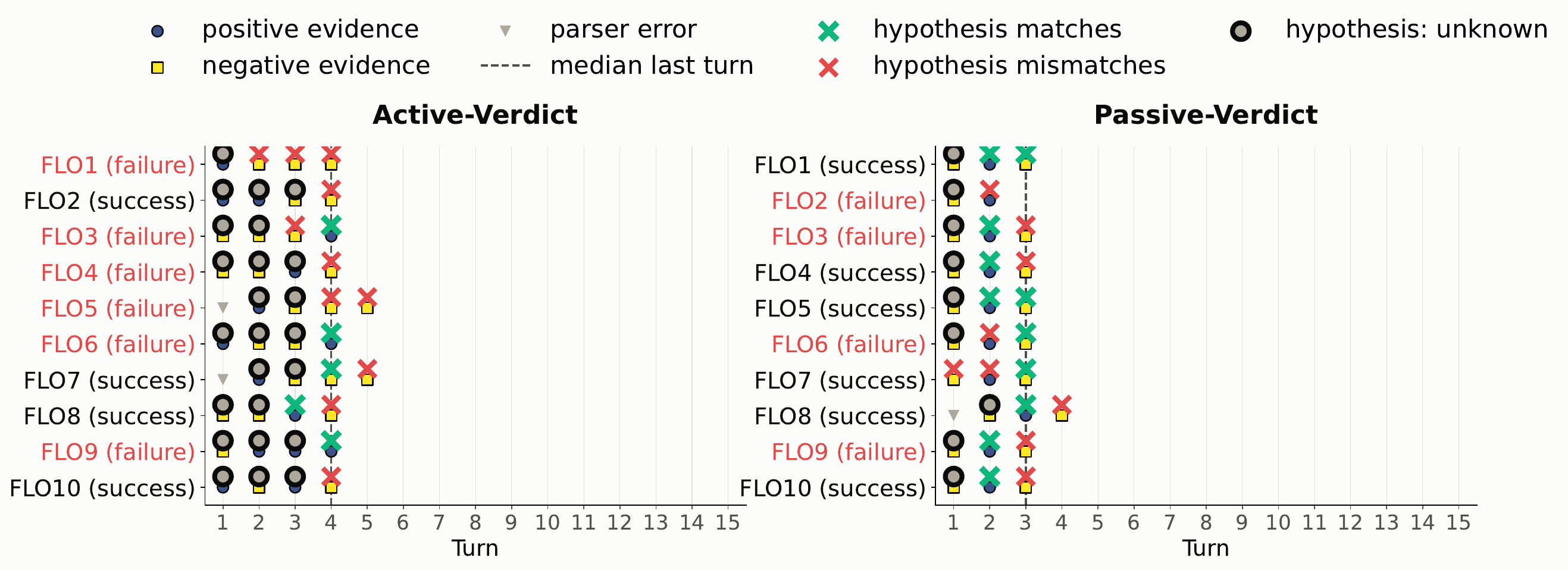}
  \caption{Negative evidence impact on task success across verdict modes 
  for GPT5.4 in Logic domain.
  }
  \label{fig:negative-logic-gpt54}
  \vspace{-0.3cm}  
\end{figure*}

\begin{figure*}[t]
\centering
  \includegraphics[width=\textwidth]{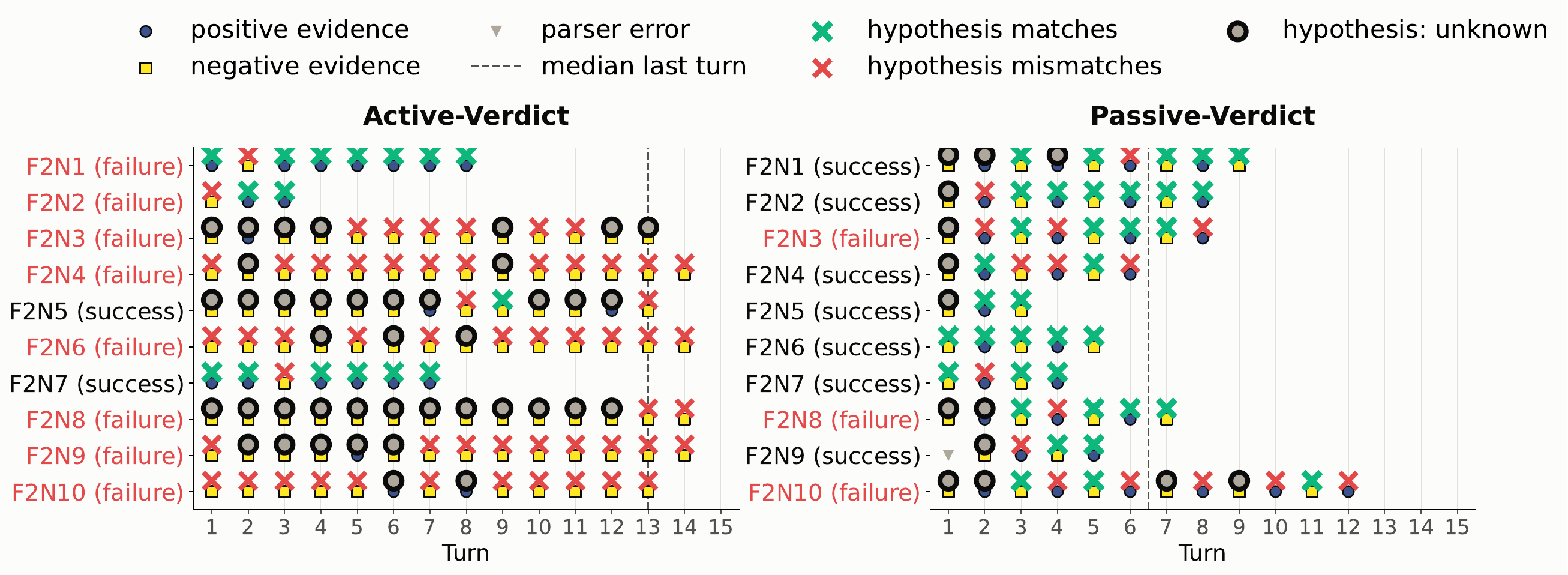}
  \caption{Negative evidence impact on task success across verdict modes 
  for GPT5.4 in Number Pairs domain.
  }
  \label{fig:negative-twonumbers-gpt54}
  \vspace{-0.3cm}  
\end{figure*}

\begin{figure*}[t]
\centering
  \includegraphics[width=\textwidth]{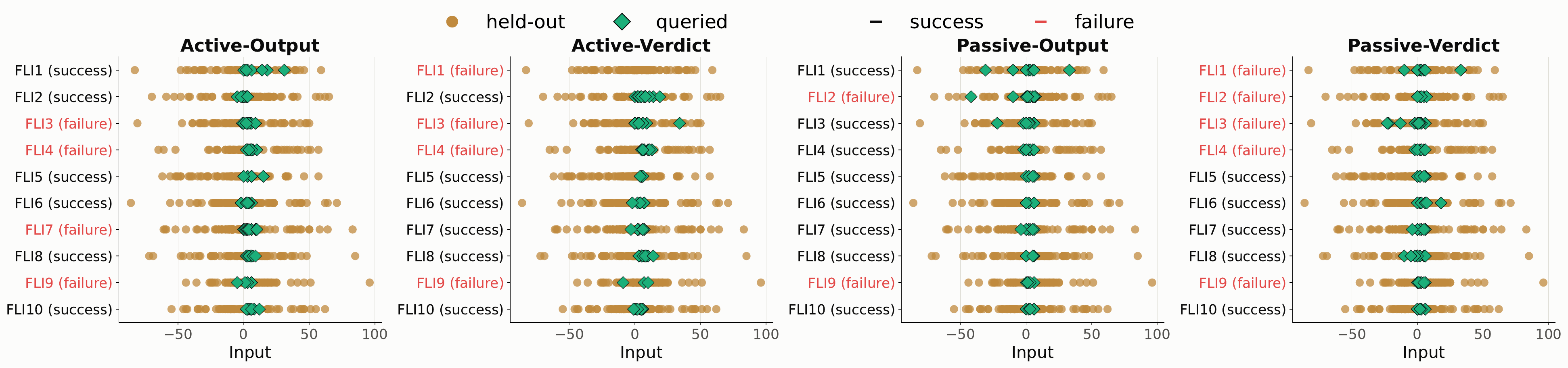}
  \caption{Distribution of input queries across interaction modes for GPT5.4 in the List domain.
  }
  \label{fig:querycoverage-list}
  \vspace{-0.3cm}  
\end{figure*}

\begin{figure*}[t]
\centering
  \includegraphics[width=\textwidth]{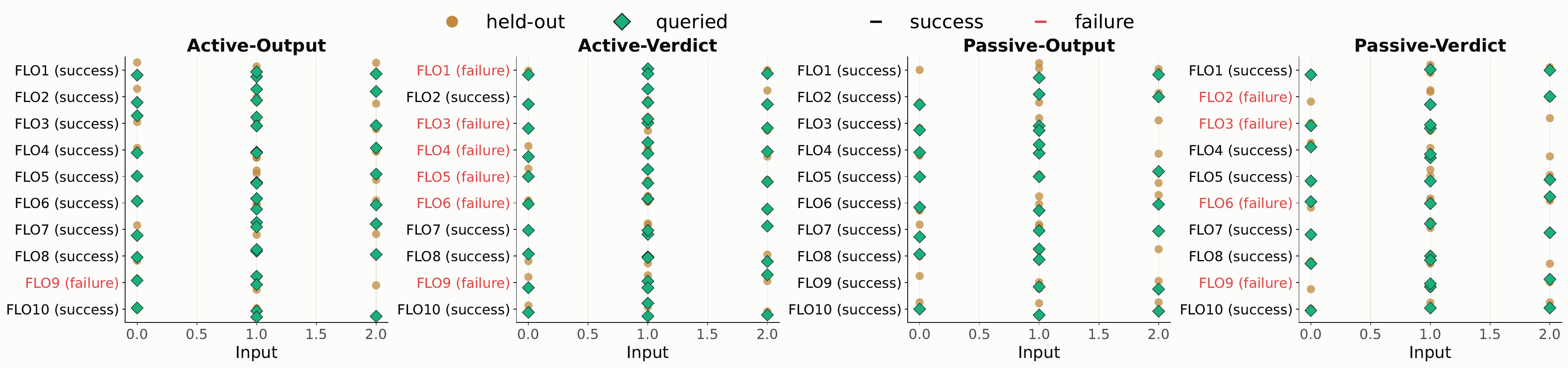}
  \caption{Distribution of input queries across interaction modes for GPT5.4 in the Logic domain.
  }
  \label{fig:querycoverage-logic}
  \vspace{-0.3cm}  
\end{figure*}

\begin{figure*}[t]
\centering
  \includegraphics[width=\textwidth]{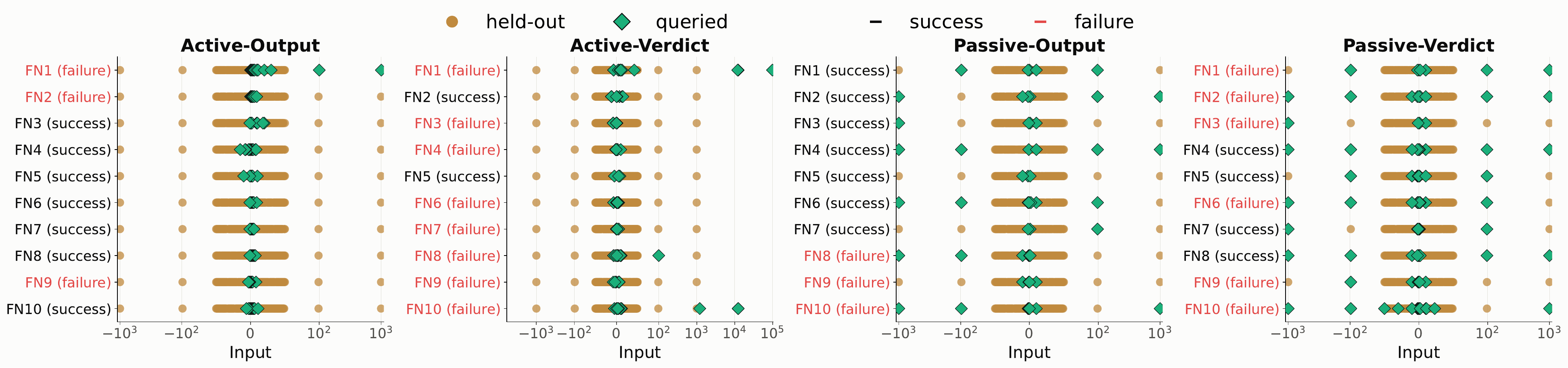}
  \caption{Distribution of input queries across interaction modes for GPT5.4 in the Numbers domain.
  }
  \label{fig:querycoverage-numbers}
  \vspace{-0.3cm}  
\end{figure*}

\begin{figure*}[t]
\centering
  \includegraphics[width=\textwidth]{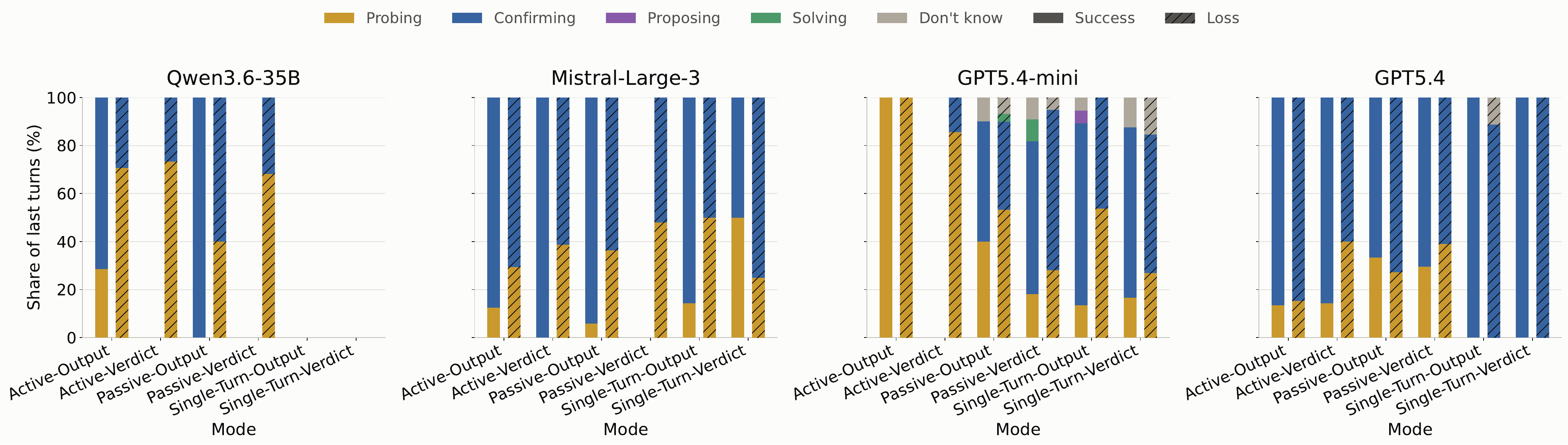}
  \caption{Interaction state at the last turn across models and interaction modes.
  }
  \label{fig:lastturn-interactionstate}
  \vspace{-0.3cm}  
\end{figure*}

\end{document}